\documentclass[sigconf]{acmart}

\usepackage[normalem]{ulem}
\usepackage{hyperref}
\usepackage{enumitem}
\usepackage{caption}
\usepackage{subcaption}

\AtBeginDocument{%
\providecommand\BibTeX{{%
\normalfont B\kern-0.5em{\scshape i\kern-0.25em b}\kern-0.8em\TeX}}}

\copyrightyear{2020}
\acmYear{2020}
\setcopyright{acmcopyright}\acmConference[MMSports'20]{3rd International Workshop on Multimedia Content Analysis in Sports}{October 16, 2020}{Seattle, WA, USA}
\acmBooktitle{3rd International Workshop on Multimedia Content Analysis in Sports (MMSports'20), October 16, 2020, Seattle, WA, USA}
\acmPrice{15.00}
\acmDOI{10.1145/3422844.3423054}
\acmISBN{978-1-4503-8149-9/20/10}

\begin{document}
\fancyhead{}

\title{Self-Supervised Small Soccer Player Detection and Tracking}

\author{Samuel Hurault}
\email{samuel.hurault@ens-paris-saclay.fr}
\affiliation{%
  \institution{ENS Paris-Saclay}
  \streetaddress{}
  \city{Cachan}
  \country{France}}

\author{Coloma Ballester}
\email{coloma.ballester@upf.edu}
\affiliation{%
  \institution{Universitat Pompeu Fabra}
  \streetaddress{}
  \city{Barcelona}
  \country{Spain}}

\authornote{co-supervision}
  
\author{Gloria Haro}
\email{gloria.haro@upf.edu}
\affiliation{%
  \institution{Universitat Pompeu Fabra}
  \streetaddress{}
  \city{Barcelona}
  \country{Spain}}

\authornotemark[1]


\begin{abstract}

In a soccer game, the information provided by detecting and tracking brings crucial clues to further analyze and understand some tactical aspects of the game, including individual and team actions. 
State-of-the-art tracking algorithms achieve impressive results in scenarios on which they have been trained for, but they fail in challenging ones such as soccer games.  This is frequently due to the player small relative size and the similar appearance among players of the same team.
Although a straightforward solution would be to retrain these models by using a more specific dataset, the lack of such publicly available annotated datasets entails searching for other effective solutions.
In this work, we propose a self-supervised pipeline which is able to detect and track low-resolution soccer players under different recording conditions without any need of ground-truth data. 
Extensive quantitative and qualitative experimental results are presented evaluating its performance. We also present a comparison to several state-of-the-art methods showing that both the proposed detector and the proposed tracker achieve top-tier results, in particular in the presence of small players. \emph{Code available at \url{https://github.com/samuro95/Self-Supervised-Small-Soccer-Player-Detection-Tracking}}.

\end{abstract}




\begin{teaserfigure}
\centering
  \includegraphics[width=0.8\textwidth]{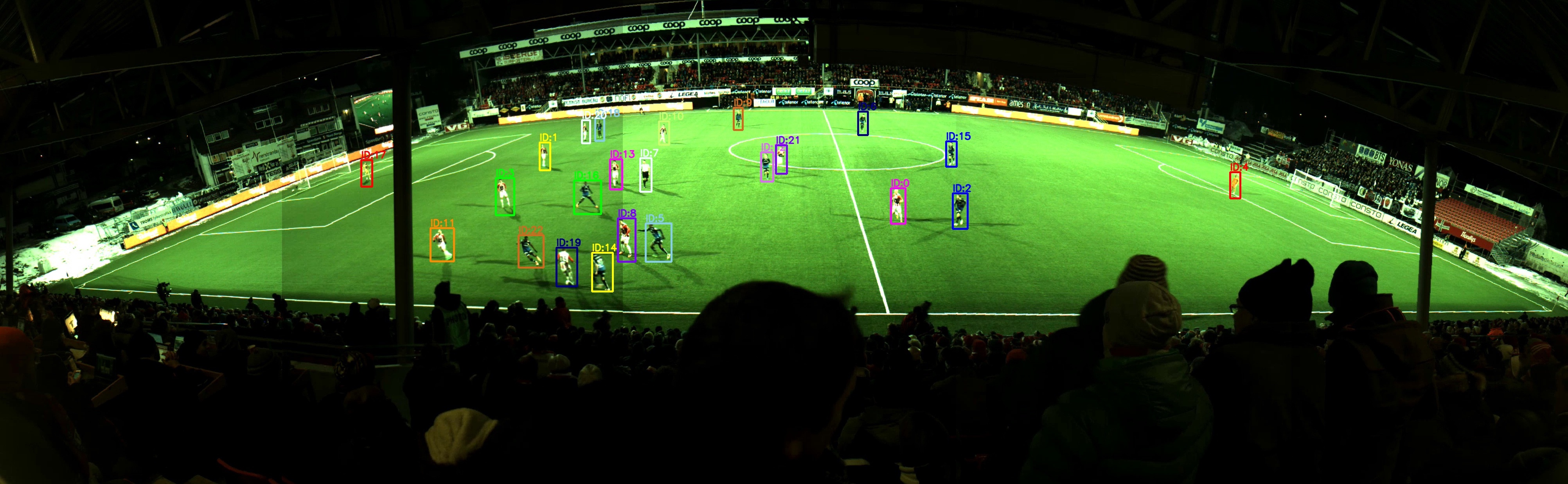}
  \caption{Extracted frame from our proposed tracking method applied to a wide-angle view soccer video sequence.}
  \Description{}
  \label{fig:teaser}
\end{teaserfigure}

\maketitle

\section{Introduction}
Tracking data is nowadays a crucial information for professional sport teams. There is a growing demand for automatic statistic collection and analysis of soccer matches. Along with event data, player tracking data is a fundamental information to build automatic action understanding.  

In this work, we aim to construct a simple and efficient soccer player tracking system that can be quickly and easily used after capturing a game. It is able to detect and track even in challenging scenarios such as the frame of the video shown in Figure~\ref{fig:teaser}. In such a camera setting, where a single wide-angle fixed camera is placed roughly at the center of the field, all players are visible in the video, which makes possible to track all of them during the whole game with just a single camera. However, this kind of view renders very small and low-resolution players on the image, making both their detection and their individual identification for tracking highly challenging. An alternative could be to use 4K images, as in \cite{4K_cropping}, in order to obtain higher-resolution players, but this solution is not always available and it leads to very time-consuming algorithms.

In order to realize tracking in videos, recent proposals use a frame-wise detection in a first step followed by an association step that links detections in consecutive frames. The effectiveness of those methods heavily relies on the accuracy of the object detector. Our proposed method follows this same strategy while being robust to the aforementioned drawback: in a first stage we  
train, without any manual annotations, a  soccer player  detector capable of recognising small players and then we establish associations among the previous detections, again following a self-supervised approach.

Numerous Deep Neural Networks (DNN) architectures \cite{Faster-RCNN,yolo,ssd} for object detection have been widely studied in the literature and achieve remarkable results. A first straightforward approach to detect players would be to use a pretrained object detection model, trained on a general dataset, like COCO~\cite{coco}, and to extract the human-class detections. We show in the supplementary material an application of a Faster R-CNN ResNet-50 FPN model pretrained on COCO \cite{coco}, applied to a typical image of a soccer game. 
When these models are applied to detect players in usual soccer games videos, their performance is strongly affected. The reasons are numerous and include soccer player singular appearances, small relative size of far away players, presence of numerous non-player humans (like spectators, coaches), motion blur or strong occlusions.

For all these reasons,  DNN models for soccer player detection  are usually supervised, i.e., trained on ground-truth player positions. However, the manual annotation of bounding boxes is costly and time-consuming. To the best of our knowledge, no dataset of annotated soccer games, in various illumination conditions and scenarios 
has been made publicly available. Thus, we opt for a self-supervised strategy.

Our goal is to automatically achieve a knowledge transfer from a pretrained general human detector to a soccer player detector, without any manual annotation on the soccer images. Our final model is able to detect very small players, avoiding other humans present on the image (like spectators or coaches). Our approach is end-to-end as it takes as input the image without any pre-processing step and outputs the player positions. Figure \ref{fig:teaser} illustrates a detection result using our proposal. 
As it can be observed, small-size players are successfully detected. Additionally, in the same figure we illustrate the tracking result using different colors for the video frame bounding boxes  corresponding to different tracking identities.
To summarize, the contributions of our work are~:\\
     $\bullet$ A soccer player detector and tracker for wide-angle video games.\\
     $\bullet$ A training method that does not require ground-truth data and learns a detector and a tracker that are accurate and robust to small players.\\
     $\bullet$ An ablation study with a thorough quantitative analysis of all the steps of the proposed training procedure and its benefits for the final proposed detector and tracker, including a comparison with state-of-the-art methods.


The outline of the paper is as follows. Section \ref{sec:relwork} reviews the related work. Section \ref{sec:method} presents the proposed method for detection and tracking, with the main details on the model and architecture (more detailed information can be found in the supplementary material). An experimental evaluation of the proposed method together with a comparison to previous works is given in Section \ref{sec:experiments}. Finally, conclusions are drawn in Section \ref{sect:conclusions}.

\section{Related work}\label{sec:relwork}
In this section, we survey the main proposed methods and deep architectures for object detection and tracking, in particular for small-size objects, and their applications to sport analysis.

\emph{Object detection with deep learning.}
CNNs have considerably improved the accuracy on object detection. Two families of detectors are currently popular: First, region proposal based detectors R-CNN \cite{RCNN}, Fast R-CNN \cite{fast-rcnn}, Faster R-CNN \cite{Faster-RCNN} and, second, detectors that directly predict boxes for an image in one step such as YOLO~\cite{yolo} and SSD~\cite{ssd}.  The goal of region proposal based detectors is to find regions of interest in the image that might contain an object and extract CNN features to classify them. Faster R-CNN \cite{Faster-RCNN} merges region proposal and classification phases into a single and end-to-end model that is faster and more accurate than the previous one. Feature pyramid networks (FPN) \cite{FPN} were built on top of R-CNN systems to enhance their capacity to detect objects of varying sizes. They exploit the inherent multi-scale, pyramidal hierarchy of the CNN backbone of these networks to construct a pyramid of features. R-CNN sub-networks are then built on these semantically-strong multi-scale backbones. However, FPN models still show limited performances for detecting small objects.

\emph{Self-supervised domain adaptation for Object Detection.}
Some works have developed strategies to tackle the absence of annotated data for object detection in a particular domain. For example, some approaches  use synthetically generated  training images \cite{DA_gan1,DA_gan2}, which are not competitive in the context of soccer. Similar to us, \cite{arthus} uses a distillation framework to transfer the knowledge of a teacher network pretrained on the source domain to a student on the target domain. This simple approach has the advantage of not requiring any annotated data in the target domain.  \cite{DA_tracking} proposes  to improve the distillation by adding additional temporal consistency information obtained via object tracking. 

\emph{Small object detection.}
Several methods based on generating new and 
challenging data have recently been proposed to enhance the capacity to detect small objects. 
Some works propose small object data augmentation: 
\cite{small_obj_aug} copy-pastes small objects and \cite{ssd} directly performs a zoom out operation on training images. Other methods leverage generative adversarial strategies 
to generate either super-resolved feature representations \cite{li2017perceptual} or super-resolved image patches \cite{bai2018sod}. 
Besides, several works (e.g. \cite{context1,context2}) use  context information surrounding the object proposal to improve their 
understanding. 

\emph{Soccer player detection.}
Many non deep learning methods have been developed to detect soccer players 
\cite{,review_detection_soccer_2,manafifard2017survey}. Classical approaches can be divided into two
groups: blob-based approaches relying on background subtraction \cite{blob_detection,background_substraction} and feature-based classification methods \cite{feature_detection_1,feature_detection_2}. Beyond,  deep learning approaches have exhibited great capacity in render more robust methods in many contexts.  Probably due to the lack of publicly available annotated data, most of the sport player detection algorithms use the mentioned pretrained deep learning human detector models without fine-tuning on the specific sport scenario \cite{simple_detection_deep,WhoHastheBall,4K_cropping}. Some methods like \cite{cascaded_player,FootAndBall} devise specific CNN architectures for player detection. To train their models, \cite{cascaded_player} manually annotates images from two soccer games and  \cite{FootAndBall} uses these annotations as well as the annotated game given by \cite{issia}.  Their models are trained and tested, respectively, on two and three games only. This is insufficient in order to capture the visual variability of soccer games with e.g.  various team color, player sizes, different stadiums or weather conditions. 
A method for detecting small players in soccer applied on multi-view videos is proposed by \cite{4K_cropping}. Notice, that we focus on a more challenging scenario for tracking with single-camera views. As mentioned, \cite{arthus} performs online semantic segmentation without requiring any manual annotation. On each specific game, they train a fast student network with data annotated by a slower pretrained teacher. This strategy can achieve good results but it adds complexity and the need of GPU resources to fine-tune on each specific game. Also, they evaluate their model on a standard TV broadcast game without focusing on possible small player failure cases. 

\emph{Multi-object tracking.}
Multiple Object Tracking (MOT) refers to the task of estimating trajectories over time of multiple objects
(see \cite{ciaparrone2020deep,leal2017tracking,manafifard2017survey} for a detailed review). As previously mentioned, one of the most popular strategies for MOT is tracking by detection which mainly consists of a first step where the objects are detected followed by an identification step where the same object is matched through time and joined in individual trajectories. For instance, \cite{bergmann2019tracking} proposes to adapt a state-of-the-art detector network into a tracker under the assumption  that the target objects slightly move between consecutive frames. Then, a  re-identification step trained on ground-truth tracking data is added to improve the results.  Other approaches use graph-based algorithms for the tracking step (e.g., \cite{yang2012online,leal2016learning,braso2020learning}).
The so-called single-shot methods aiming to jointly detect and identify the targets have recently gained a lot of attention \cite{feichtenhofer2017detect,wang2019towards,zhan2020simple,braso2020learning}. In \cite{wang2019towards} a single network integrates both the detector and an embedding model for identification. Similarly,
\cite{feichtenhofer2017detect} processes two consecutive frames and generates two-frame tracklets that are afterwards combined into multi-frame tracks.
%
A group of MOT methods leverage pose estimation of humans for tracking purposes \cite{ning2020lighttrack,girdhar2018detect,xiao2018simple,ran2019robust}.
%
However, tracking strategies based on human pose exhibit difficulties as they struggle in correctly estimating pose for low-resolution persons such as the small players that usually appear in real wide-angle soccer videos, as tackled in this work.




\section{Method}\label{sec:method}
We propose a self-supervised soccer player detector and tracker that can be applied to wide-angle video games since it is robust to small players. Our method is specialized in soccer and does not require any labeled data. Our pipeline follows two steps : first a soccer player detection and then a tracking based on the detection results. In Section \ref{detection_section}, we describe the detection framework which consists of the successive training of a teacher and  a student for domain adaptation and knowledge distillation. Then, in Section \ref{sec:soccer-player-tracking-framework}, we describe the tracking framework which uses the very same detector to build player trajectories using spatial and visual consistency. 

\subsection{Soccer player detection}
\label{detection_section}

\begin{figure}
  \includegraphics[width=\linewidth]{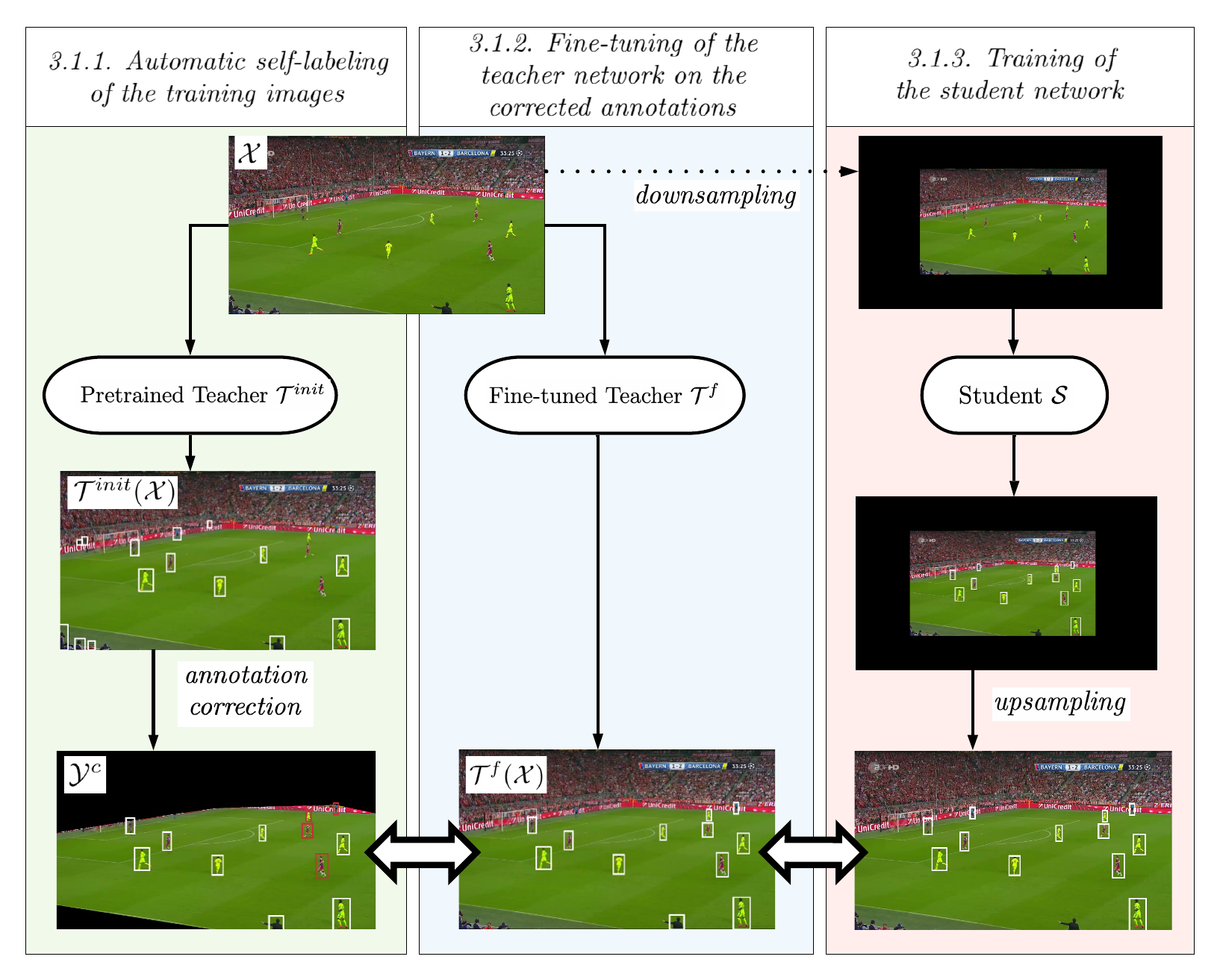}
  \caption{Illustration of our soccer player detection method. Each column corresponds to one step of the proposed method explained in Section \ref{detection_section}.}
  \Description{}
  \label{fig:pipepile}
\end{figure}

Our first goal is to train an accurate player detector. 
To do so, we specialize an object detection model, trained on a general dataset, in the detection of soccer players, without any further manual annotations.  From a set of non-annotated training soccer images, denoted by $\mathcal{X}$, we adopt the following three-step training process also illustrated in  Figure~\ref{fig:pipepile}~: 
\begin{enumerate}
    \item Label the set of training images $\mathcal{X}$ with a pretrained human detector denoted as the initial teacher network  $\mathcal{T}^{init}$. This set of initial noisy annotations $\mathcal{T}^{init}(\mathcal{X})$, are automatically corrected  by suppressing False Positives and adding False Negatives resulting in a  new set of labels  $\mathcal{Y}^c$.
    \item Fine-tune the initial teacher network on these corrected annotations  $\mathcal{Y}^c$  in order to get the final teacher network $\mathcal{T}^{f}$ and its corresponding set of annotations $\mathcal{T}^{f}(\mathcal{X})$.
    \item Distil the knowledge of the teacher $\mathcal{T}^{f}$ to a smaller and faster student network  $\mathcal{S}$ which is our final accurate detector.
\end{enumerate}


The baseline object detector we use is the well-known Faster-RCNN \cite{Faster-RCNN} network with Feature Pyramid Network (FPN) \cite{FPN}. It is widely used in the literature due to its robustness to various object sizes while keeping low inference time. We use, respectively, ResNet50 and ResNet18 backbones for the teacher and student networks. We want our whole domain adaptation process to stay general so that it can be easily applied to any other object detection model. Therefore, we do not apply hard changes to the original FPN Faster-RCNN architecture. We adapt some  parameters to make the model more sensitive to small players while keeping a reasonably low inference time. More details about the parameter choices can be found in Section \ref{sec:mmplementation} and in the supplementary material. 

\subsubsection{Automatic self-labeling of the training images}
\label{sec:labeling}

Our training images $\mathcal{X}$ are unlabeled TV broadcast soccer videos taken from a subset of the SoccerNet dataset \cite{SoccerNet}. More details about the dataset can be found in Section \ref{sec:traininngdataset}. We extract from $\mathcal{T}^{init}(\mathcal{X})$ the human annotations with a confidence higher than $\sigma=0.8$. These annotations are then post-processed by the following two steps.

\paragraph{Field detection}
Most of the False Positives consist of detections of supporters and coaches. We use a combination of field and line detection to avoid them. A first field mask is computed through green filtering and contour detection. To detect the line we use the pix2pix line extraction model of \cite{line_pix2pix}. 
An example can be found in the bottom-left corner of Figure \ref{fig:pipepile}. 
Additional details and results can be found in the supplementary material. Finally, all the detected bounding boxes with a proportion of the area in the field mask lower than $0.3$ are discarded. Note that in practice this post-processing is also used to correct the $\mathcal{T}^{f}(\mathcal{X})$ annotations.

\paragraph{Adding missed players to the training data}
We also add missed detections to the training data through a blob detection strategy. This is realized via green filtering, contour detection and human detection in the regions of the contours with appropriate size. Detailed information can be found in the supplementary material.  

\subsubsection{Fine-tuning of the teacher network on the corrected annotations}
We perform a fine-tuning of the teacher $\mathcal{T}^{init}$ network on the corrected annotations  $\mathcal{Y}^{c}$. For training, we use the same loss as in \cite{Faster-RCNN}. Let us denote this detection loss by $\mathcal{L}_{detection}$. 
This first retraining step will enable the big teacher network to first adapt to the new domain before transmitting the knowledge to a smaller student network. As shown in Section \ref{sec:experiments}, this procedure improves the final student accuracy.

\subsubsection{Training of the student network} \label{sec:training_student}

\paragraph{Student $\mathcal{S}$ architecture} We propose one small modification of the student $\mathcal{S}$ architecture    to make the model consider the context surrounding a detection before assigning it a bounding box. More precisely, we follow the idea proposed by \cite{context1} and incorporate the context region enclosing each proposal in the decision process. This is done by concatenating to the feature maps of an extracted region of interest bounding box, the feature maps of this same box enlarged by a factor of $4$. 
As explained in \cite{context1}, adding the contextual regions is likely to help the detection of small objects.

\paragraph{Down-scaling of the training images}
As explained in the introduction, the goal of our object detector is to be robust to small players like the ones that appear in Figure~\ref{fig:teaser}.
On this kind of views, the entire field is constantly visible in the image. 
However, our training data (e.g.~ Figure~\ref{fig:pipepile}) is composed of TV broadcast images that are often zoomed on the current game action. The players thus appear substantively bigger on the screen. We also want our detector to be robust to varying player sizes. To do so, we perform data augmentation by re-scaling each training image by a random factor chosen between a given value and $1$. Let us denote this value by $min\_scale$, which verifies $0<min\_scale\leq 1$. We add zero-padding to the resulting images to make them match their original size. 

To train the final student network $\mathcal{S}$, we extract from $\mathcal{T}^{f}(\mathcal{X})$ the annotations with a confidence higher than $\sigma=0.8$, post-process them by field detection and finally randomly down-scale the input image with $min\_scale$ equal to $0.1$.

\subsection{Soccer player tracking framework}
\label{sec:soccer-player-tracking-framework}

We propose a tracking approach to associate the previous detections between consecutive frames. Our method is unsupervised, accurate for small players, simple and fast.  It is only based on our player detector model.   
Given a video sequence $\{u_t\}_t$ where $t$ denotes time, we define a player trajectory as a finite list of ordered player bounding boxes $T_k = \{b_{t_1}^{k},b_{t_2}^{k},...\}$. Here we assume that $0\leq t_1<t_2<...$. 
At each frame $u_t$, we compute the set of potential bounding boxes $\mathcal{B}_t =\{b_{t}^{k_1},b_{t}^{k_2},...\}$ by running our player detector $\mathcal{S}$ and gathering all the annotations $\mathcal{S}(u_t)$ with a confidence higher than $\sigma_{track}>0$. To extract the player trajectories from these bounding boxes, we propose to consider two successive steps, explained in the following subsections.

\subsubsection{Spatial consistency association}
\label{spatial_consistency}

We first rely on spatial consistency, i.e., if two bounding boxes from the current and the previous frames are adjacent we consider them to belong to the same target. Formally, at $t>0$, for each player bounding box $b_{t}^{i}$ in $\mathcal{B}_t$ if there is a unique bounding box $b_{t-1}^{j}$ in $\mathcal{B}_{t-1}$ such that the intersection-over-union between these two bounding boxes satisfies $IOU(b_{t}^{i},b_{t-1}^{j})>\tau_{iou}$, these detections are associated to the same track. In case $b_{t-1}^{j}$ is associated to two different players in $\mathcal{B}_t$, these associations are discarded. Once a bounding box at frame $t$ is associated to a track, it is removed from $\mathcal{B}_t$.

However, as explained in \cite{ning2020lighttrack} and \cite{bergmann2019tracking}, the above criterion is based on the assumption that the tracked target has significant overlap between two successive frames. This assumption is not satisfied in some cases such as in low frame rate, fast player movements, or intense camera shifts. Furthermore, in the presence of crowded scenarios with partial occlusions between players, the IOU measure is not satisfactory. For these reasons, we consider a complementary visual consistency measure. 

\subsubsection{Re-identification and visual consistency}
\label{visual_consistency}

The  players from frame $u_{t-1}$ non directly associated with those from frame $u_t$ with the previous criterion are first considered to be lost. We store killed (deactivated) tracks for a fixed number of $N_{reID}$ frames. 
We then use a visual embedding extracted from the player detector network to compare the visual appearances of the  remaining bounding boxes in $\mathcal{B}_t$ with the deactivated tracks. Using an additional ResNet50 like \cite{bergmann2019tracking} would be time consuming, so we prefer to use the player detector model to extract visual information. The advantage is that this information is already freely available for each detection. 

We propose to keep the original player detector $\mathcal{S}$ architecture and to directly extract the embedding features given by ROI pooling of the region proposals. However, a fine-tuning of the network is needed to make the individual features  discriminative enough between players. Let us remark that we found that modifying the architecture by adding in parallel to the ROI head, a Siamese "Tracking head" (\cite{shuai2020multi}) did not improve the quality of the final extracted embedding. Our proposed method remains simple and enables to unify detection and association, without any change in the architecture of the detector.

In principle, this fine-tuning would require tracking ground-truth data. Due to the absence of such data in soccer, we propose to use the following unsupervised approach: 
\begin{itemize}
    \item[(1)] Use annotations $\mathcal{B}_t =\{b_{t}^{k_1},b_{t}^{k_2},...\}$ given by the teacher $\mathcal{T}^{f}$.
    \item[(2)] Generate trajectories $\{T_k\}$ only with the spatial consistency association method explained in Section  \ref{spatial_consistency}.
    \item[(3)] Keep only long enough trajectories, i.e. those where $|T_k|>5$ frames.
\end{itemize}
We fine-tune the student network  $\mathcal{S}$ on this data with a triplet loss, using the \textit{batch hard} strategy introduced in \cite{hermans2017defense}. Given a small training sub-sequence of $N$ consecutive frames, we randomly sample $K$ player tracks, and then randomly sample $T$ detections for each track. This ensemble of detections forms a batch of $KT$ samples. 
We define the triplet loss over this batch as :
\begin{equation}\label{eq:tripletloss}
\begin{split}
\mathcal{L}_{triplet}(\theta) = \sum_{k=1}^{K} \sum_{t=1}^{T} \left[ m + \max_{u=1...T} || f_{\theta}(b_{t}^{k}) - f_{\theta}(b_{u}^{k}) ||_2 \right.\\
\left.- \min_{i=1...K, u=1...T, i \ne k} || f_{\theta}(b_{t}^{k}) - f_{\theta}(b_{u}^{i}) ||_2 \right]_{+}
\end{split}
\end{equation}
where $m>0$ denotes the margin (in practice, we use $m=2$), $\|\tilde{f}\|_2=<\tilde{f},\tilde{f}>^{1/2}$, for $\tilde{f}\in\mathbb{R}^M$, and $[x]_+$ denotes, as usual, the maximum between a real value $x$ and $0$. $f_{\theta}$ takes a bounding box as input and outputs the resulting identification vector. $\theta$ denotes the weights of the backbone and ROI pooling layers. 

In parallel, we need  to back-propagate the detection loss in order not to lose  the ability of the network to detect players. The detection loss, $\mathcal{L}_{detection}$, is computed with  the teacher network annotations (see Section \ref{detection_section}), with respect to the whole network parameters $\tilde{\theta}$ and it is summed over the $N$ frames of the working sub-sequence. The final training loss is $\mathcal{L}(\tilde{\theta}) = \mathcal{L}_{detection}(\tilde{\theta}) + \mathcal{L}_{triplet}(\theta)$. 

\paragraph{Tracking inference}
For inference, we match the remaining elements in $\mathcal{B}_{t}$ with the deactivated tracks by the Hungarian  graph bipartite matching algorithm \cite{kuhn1955hungarian}. The distance between $b_{t}^{i}$ and a killed track $T_k = \{b_{t_1}^{k},b_{t_2}^{k},...\}$ is averaged over the last elements of the track. We define the single association cost between $b_{t}^{i}$ and one player $b_{t_j}^{k}$ of the track as a combination of visual and spatial distances : 
\begin{equation} \label{eq:cost}
C(b_{t}^{i},b_{t_j}^{k}) = \alpha D_{visual}(b_{t}^{i},b_{t_j}^{k}) + (1-\alpha) D_{spatial}(b_{t}^{i},b_{t_j}^{k})
\end{equation} 
where $0\leq\alpha\leq 1$, 
\begin{equation}
D_{visual}(b_{t}^{i},b_{t_j}^{k}) = ||f_{\theta}(b_{t}^{i}) - f_{\theta}(b_{t_j}^{k})||_{2},
\end{equation} 
\begin{equation}
D_{spatial}(b_{t}^{i},b_{t_j}^{k}) = ||x_{c}(b_{t}^{i})-x_{c}(b_{t_j}^{k})||_{2},
\end{equation} 
and $x_{c}(b)$ denotes the center point of a bounding box $b$. 
A player-track association is established under two additional conditions on the visual and spatial distances : $D_{spatial} < D_{spatial\_max}$ and $D_{visual} < D_{visual\_max}$.



\section{Experiments} \label{sec:experiments}

\subsection{Training data}\label{sec:traininngdataset}

We train our method on a subset of SoccerNet dataset \cite{SoccerNet}. In particular, we extract only wide-angle sequences where the game is filmed with a wide angle of a lateral view from a fixed camera placed, roughly, at the center of the lateral tribune. Examples of these images are shown Figure~\ref{fig:pipepile} and in the supplementary material. It includes $2321$ randomly extracted $5$ seconds sequences from $50$ unlabeled TV broadcast soccer games of resolution $720 \times 1280$ and frame rate $25$ fps. In total, it comprises $290125$ images. Different stadiums, illumination conditions, teams, and viewpoints are included.
\begin{figure}
  \includegraphics[width=0.7\linewidth]{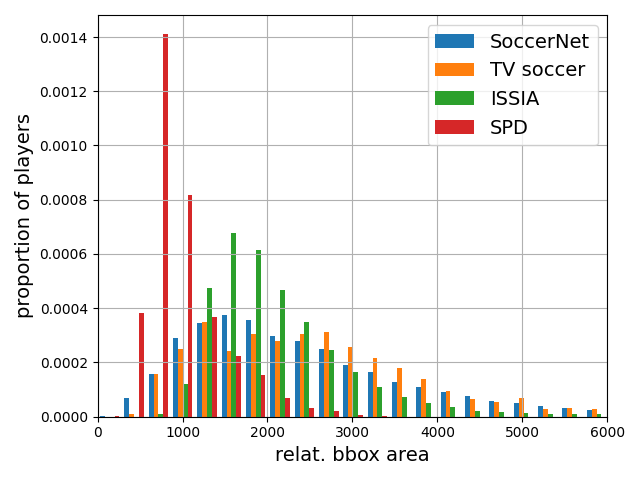}
  \caption{Histograms of the  relative bounding box areas of players for each dataset. The SoccerNet statistics are computed with annotations given  by the teacher network.}
  \label{fig:bbox_histogram}
\end{figure}
\begin{table}
  \caption{Statistics on different  detection datasets used. "bbox area" refers to the average area of the ground-truth player bounding boxes. "relat. area" is the average bounding box area divided by the whole image area. In SoccerNet we provide estimations since there is no ground truth annotation.}
  \label{tab:eval_data_stats}
    \setlength{\tabcolsep}{4pt} 
\renewcommand{\arraystretch}{0.7} 
  \begin{tabular}{c|c|c|c|c}
    \toprule
    Dataset & \# images & image size & bbox area & relat. area ($\times10^6$) \\
    \midrule
    SoccerNet & $290125$ & $720 \times 1280$ & $\approx 2200$ & $\approx 2300$ \\
    $TVsoccer$ & $250$ & $720 \times 1280$  & $2580$ & $2300$ \\
    $ISSIA$ & $2600$ & $1080 \times 1920$ & $4420$ & $2140$ \\
    $SPD$ & $550$ & $720 \times 1280$ & $921$ & $950$ \\
  \bottomrule
\end{tabular}
\end{table}

\subsection{Testing data}\label{sec:testing}
In order to quantitatively evaluate our detection and tracking results, we need to assess them on ground-truth annotated data. For detection evaluation, the position bounding boxes of all the players (and referees) in the field are needed. We use three different player position datasets for evaluation:
\begin{itemize}
    \item "$TVsoccer$" : we manually annotated $250$ images with wide views extracted from different TV broadcasted games of \cite{Yu2018ComprehensiveDO}. 
    \item "$ISSIA$" :  $2600$ frames from ISSIA-CNR (annotated) dataset \cite{issia}. We only perform our evaluation on camera $3$ because it is the one placed on the center of the field and thus its view contains most of the players. 
    \item "$SPD$" :  $550$ wide view annotated frames extracted from \cite{cascaded_player}.
\end{itemize}
Various example images from these datasets are shown in the supplementary material.
Additionally, statistics about these datasets can be found in Table \ref{tab:eval_data_stats}, in particular comparing the different sizes of the players. The fourth column  shows the average player bounding box sizes. Moreover, as the first pre-processing step of the player detector is to rescale the input image to a fixed image 
size, we also display in the fifth column 
the relative area, i.e., the average bouding box area divided by the whole image area. 
Figure \ref{fig:bbox_histogram} also plots the histograms of the relative  bounding box areas in each dataset.  Let us notice that the "$TVsoccer$" and "$ISSIA$" 
have similar player sizes to the SoccerNet training data, and   that "$SPD$" contains  smaller players than the other datasets. In order to fit the average size of training data with the one of "$SPD$", a rescaling of the training images with a scaling factor of approximately $0.6$ would be needed.

To evaluate the tracking performance, an individual identity annotation is needed. We use the $ISSIA$-CNR dataset \cite{issia} and test on $5$ different sequences of $10$ seconds at $25$ fps from camera $3$. 

We also qualitatively evaluate our proposed detection and tracking methods on sequences extracted from \cite{panorama_dataset} (an example can be found in Figure \ref{fig:teaser}) where really small players usually appear.

\subsection{Evaluation metrics}
For evaluation of the player detection model, we use the standard Average Precision (AP) measure defined by the PASCAL VOC object detection challenge \cite{padilla2020survey}, at IOU threshold $0.5$. This measure evaluates the area under the Precision-Recall curve. Therefore, it determines the capacity of the detector at any score confidence threshold. However, while annotating data, we also need to evaluate the performance of the annotation at a fixed score detection threshold. In this case we will use the F1 score. It is a metric that combines recall and precision by taking their harmonic mean, 
    $F1 = \frac{precision\cdot recall}{precision + recall}$.

To measure the performance of a tracker, we focus on the Multiple Object Tracking Accuracy (MOTA) \cite{kasturi2008framework} and
ID F1 Score (IDF1) \cite{ristani2016performance}. 
MOTA and IDF1 are meaningful combinations of 
the number of false positives and negatives of the detector, the total number of identity switches, the mostly tracked and mostly lost metrics \cite{ristani2016performance}.

\subsection{Implementation details}
 \label{sec:mmplementation}

We conduct all our experiments on a GPU NVIDIA TITAN X. 
To train and evaluate the FPN-FasterRCNN models, we use as baseline the native PyTorch implementation\footnote{\url{https://github.com/pytorch/vision/blob/master/torchvision/models/detection/faster_rcnn.py}}. Most of the native parameters are unchanged from the implementation of \cite{FPN} and  \cite{Faster-RCNN}. The images are scaled to a fixed input size of $768\times1344$ as a first preprocessing step. The native NMS is replaced by Soft-NMS, as proposed by \cite{bodla2017soft}.
We use pretrained models provided by torchvision, specifically $\mathcal{T}^{init}$ on COCO 
\cite{coco} and initialization of $\mathcal{S}$ 
on ImageNet 
\cite{imagenet_cvpr09}.  
The rest of the parameters used for detection and tracking are given in the supplementary material. 

\subsection{Ablation study}
\label{sec:ablation}

\subsubsection{Player detection results}

\paragraph{Automatic labeling of the training images}

We first evaluate the proposed pre-labeling of the SoccerNet training images, described in Section  \ref{sec:labeling}. Let us remind that we first keep the pretrained   
network annotations $\mathcal{T}^{init}(\mathcal{X})$ with confidence bigger than $0.8$. We then automatically correct them with field segmentation and player blob detection to get $\mathcal{Y}^c$.  To understand its effect, we provide the statistics of the generated labels with the different methods in Table~\ref{tab:stat_labeling}. With field segmentation, approximately $5\%$ of the original annotations are removed and, with blob detection, another $5\%$ are added. It  mainly corresponds to, respectively, a reduction in false positives and an increase in true positives. 
As it can be observed in the second column of  Table~\ref{tab:stat_labeling}, the  added players are, in average, smaller than the original ones. This observation supports our claim that a pretrained detector shows poor performance for small objects. We also show in Table~\ref{tab:stat_labeling_F1} the F1-score performances, on the three testing datasets, of the models and techniques used to automatically annotate the training set.  

\begin{table}
  \caption{Statistics of the first automatic annotation process}
  \label{tab:stat_labeling}
  \begin{tabular}{c|c|c}
    \toprule
    Method & \# detections & avg bbox area \\
    \midrule
    Pretrained object detector & $14.2$/image & $2390$ \\
    + Field segmentation & $- 5.2\%$ & - \\
    + Blob detection & $+ 5.1\%$  & $1770$\\
  \bottomrule
\end{tabular}
\end{table}

\begin{table}
  \caption{F1-score of automatic annotation processes.}
  \label{tab:stat_labeling_F1}
    \setlength{\tabcolsep}{4pt} 
\renewcommand{\arraystretch}{0.75} 
  \begin{tabular}{c|c|c|c}
    \toprule
    Dataset & $TVsoccer$ & $ISSIA$ & $SPD$ \\
    \midrule
    Pretrained detection model $\sigma>0.8$ & 89.0 & 91.4 & 79.0 \\
    + Field segmentation + blob detection & 90.1 & 91.8 & 82.0 \\
    \midrule
    Retrained detection model $\sigma>0.7$ & 93.1 & 93.0 & 83.5 \\
  \bottomrule
\end{tabular}
\end{table}

\paragraph{Retraining of the teacher network on the corrected annotations}

We use these corrected annotations $\mathcal{Y}^c$ to retrain the initial teacher detection model $\mathcal{T}^{init}$ and get the final teacher $\mathcal{T}^{f}$. We have tried to fine-tune the pretrained model or to train it from scratch, with only the backbone weights pretrained on ImageNet and obtained very similar results in both cases. Therefore, in order to keep consistency with the training of other models, in all the following experiments, the teacher $\mathcal{T}^f$ 
and 
the student $\mathcal{S}$ have been trained from scratch, with ResNet50 and ResNet18 backbones pretrained on ImageNet. The training process is described in Section \ref{sec:mmplementation}.

\begin{table}
  \caption{AP performances of the retrained teacher network $\mathcal{T}^{f}$ compared to the COCO pretrained version $\mathcal{T}^{init}$. The different 
  columns indicate the results for the downscaled testing dataset.}
  \label{tab:teacher_results}
  \setlength{\tabcolsep}{3.5pt} 
\renewcommand{\arraystretch}{0.6} 
  \begin{tabular}{c|c|c|c|c|c|c|c|c|c}
    \toprule
    Dataset & \multicolumn{3}{c|}{$TVsoccer$} & \multicolumn{3}{c|}{$ISSIA$} & $SPD$ \\
    \midrule
    Rescale factor & {\small$\times$}1 & {\small$\times$}0.75 & {\small$\times$}0.5 & {\small$\times$}1 & {\small$\times$}0.75 & {\small$\times$}0.5 & {\small$\times$}1 \\
    \midrule
    $\mathcal{T}^{init}$ & 89.3 & 78.2 & 78.0 & 90.3 & 86.6 & 77.8 & 82.2 \\
    $\mathcal{T}^{f}$, $min_{scale}=1$ & \bf 95.7 & 93.9 & 84.3 & \bf 97.6 & 94.9 & 86.8 & 77.1 \\
    $\mathcal{T}^{f}$, $min_{scale}=0.5$ & 95.4 & \bf 96.5 & \bf 94.0 & 97.5 & \bf 97.7 & \bf 96.0 & \bf 92.0 \\
  \bottomrule
\end{tabular}
\end{table}

The goal of this experiment is to retrain the 
teacher 
network $\mathcal{T}^{init}$ so that it annotates as accurately as possible our original SoccerNet training images.
A first model is basically retrained on the given corrected annotations. 
We have previously seen with Table~\ref{tab:stat_labeling} that the missed detections of the pretrained model were due to the small size of the targets.
Therefore, we conduct a second training using the re-scaling data augmentation method (see Section \ref{sec:training_student}) with $min\_scale = 0.5$.
Table~\ref{tab:teacher_results} evaluates the retrained models on the three evaluation datasets. Let us remind that $SPD$ contains smaller players than $TVsoccer$ and $ISSIA$. In order to simulate smaller players in these  two last datasets, we also show in the columns {\small$\times$}0.75 and {\small$\times$}0.5, respectively, the results of the model applied to down-scaled images at fixed scales of  $0.75$ and $0.5$, respectively. Smaller scales are not explored since, as it can be  observed in the histogram of  Figure~\ref{fig:bbox_histogram}, at scale $0.5$, we already match the smallest player size present in the training images. 

Results in Table~\ref{tab:teacher_results} show that the retraining on self-annotated soccer data enables to enhance, by a large margin, the capacity of our given network to detect soccer players. We successfully perform domain adaptation of the human detector to the specific context of a soccer game. Furthermore,
we show the necessity to perform down-scaling data augmentation on our TV broadcast training images in order for the detector to be robust to small players :  using random re-scaling of the training images with $min\_scale = 0.5$ enables to almost close the gap of the detector performance  between the smallest (scale $0.5$) and the biggest  (scale $1$) players without losing performance in bigger scales. 


\paragraph{Knowledge distillation to the student network}

\begin{table}
  \caption{AP performances of the student network  training. The different columns indicate the results for the downscaled testing dataset.}
  \label{tab:student_results}
    \setlength{\tabcolsep}{3.75pt} 
\renewcommand{\arraystretch}{0.7} 
  \begin{tabular}{c|c|c|c|c}
    \toprule
    Dataset & \multicolumn{4}{c}{$TVsoccer$} \\
    \midrule
    Rescale factor & {\small$\times$}1 & {\small$\times$}0.75 & {\small$\times$}0.5 & {\small$\times$}0.3 \\
    \midrule
    $\mathcal{T}^f$ 
    & 95.4 & 96.5 & 94.0 & 83.12 \\
    \midrule
    $\mathcal{S}$ trained with $\mathcal{Y}^c$, $min_{scale}=0.5$ & \bf 95.9 & 96.3 & 94.3 & 83.8  \\
    $\mathcal{S}$ trained with $\mathcal{Y}^c$, $min_{scale}=0.1$ & 94.9 & 96.3 & 95.6 & 93.5  \\
    \midrule
    $\mathcal{S}$, $min_{scale}=0.1$ & 95.1 & 96.2 & 96.1 & \bf 93.6 \\
    $\mathcal{S}$, $min_{scale}=0.1$ w. context & 95.2 & \bf 96.5 & \bf 96.4 & 93.5  \\
  \bottomrule
\end{tabular}
\medspace

    \begin{tabular}{c|c|c|c|c}
    \toprule
    Dataset & \multicolumn{4}{c}{$ISSIA$} \\
    \midrule
    Rescale factor & {\small$\times$}1 & {\small$\times$}0.75 & {\small$\times$}0.5 & {\small$\times$}0.3 \\
    \midrule
    $\mathcal{T}^f$ & 97.5 & 97.7 & 96.0 & 87.2 \\
    \midrule
     $\mathcal{S}$ trained with $\mathcal{Y}^c$, $min_{scale}=0.5$ & \bf 97.6 & 97.3 & 96.0 & 88.3\\
    $\mathcal{S}$ trained with $\mathcal{Y}^c$, $min_{scale}=0.1$ & 96.8 & 96.7 & 97.1 & 94.3 \\
    \midrule
    $\mathcal{S}$, $min_{scale}=0.1$ & 96.6 & 96.9 & 97.4 & 95.7 \\
    $\mathcal{S}$, $min_{scale}=0.1$ w. context & 97.3 & \bf 97.7 & \bf97.9 & \bf 95.9\\
  \bottomrule
\end{tabular}

\medspace

\begin{tabular}{c|c|c|c}
    \toprule
    Dataset & \multicolumn{3}{c}{$SPD$} \\
    \midrule
    Rescale factor & {\small$\times$}1 & {\small$\times$}0.75 & {\small$\times$}0.5 \\
    \midrule
    $\mathcal{T}^f$ 
    & 92.0 & 88.7 & 83.5 \\
    \midrule
     $\mathcal{S}$ trained with $\mathcal{Y}^c$, $min_{scale}=0.5$ & 96.0 & 95.1 & 91.3 \\
     $\mathcal{S}$ trained with $\mathcal{Y}^c$, $min_{scale}=0.1$ & 96.6 & 96.2 & 96.2 \\
    \midrule
    $\mathcal{S}$, $min_{scale}=0.1$  
    & 97.7 & 97.7 & 97.5 \\
    $\mathcal{S}$, $min_{scale}=0.1$ w. context & \bf 98.0 & \bf 
    97.8 & \bf 97.9 \\
  \bottomrule
\end{tabular}
\end{table}

Assessment on our final detector is now presented. We use the annotations generated by the previously retrained 
teacher $\mathcal{T}^f$ 
to train the 
student network $\mathcal{S}$. Once again, we train it from scratch, using the ResNet18 backbone pretrained on ImageNet. We follow the same training process described in Section \ref{sec:mmplementation}. We train with random down-scaling of the training images, but this time with $\min_{scale}=0.1$ in order to get a detector as robust as possible to small players. We also show in Tables \ref{tab:student_results} the results using the context feature map concatenation method introduced in Section \ref{sec:training_student}.
The AP performances of these trainings are reported, for each evaluation dataset, and for down-scaling factors \{1, 0.75, 0.5, 0.3\} in the third block of lines in each subtable of Table~\ref{tab:student_results}. $SPD$ is not evaluated at scale $0.3$ due to the already small size of the players on the original images. For comparison, we also train 
$\mathcal{S}$ 
like the previous teacher on the original corrected annotations $\mathcal{Y}^c$ (results in the second block of lines in each subtable of  Table~\ref{tab:student_results}). For the sake of comparison, we also report the performance of the teacher  in the first blocks of Table~\ref{tab:student_results}.

The following observations can be extracted~: \\
     \textit{i)} The first two lines of results in each subtable show that ResNet18 and ResNet50 backbones bring already similar results on $ISSIA$ and $TVsoccer$ when retrained on the original corrected annotations, while the ResNet18 backbone outperforms ResNet50 by a large margin on the $SPD$ dataset. We hypothesize this difference by the fact that there exists a difference in the appearance of small players present in the $SPD$ dataset and  small players  "simulated" by downscaling the SoccerNet image : as players in $SPD$ are filmed from a higher point of view than most of the shots in SoccerNet, $TVsoccer$ and $ISSIA$. It is very likely that a Resnet50 backbone, with higher complexity, slightly over-fits on the SoccerNet player appearances when the Resnet18 version, with less parameters, better generalizes to new soccer player appearances. We add additional details about this observation in the supplementary material. \\
     \textit{ii)} Using $min\_scale=0.1$ enables a substantial boost of the performance for small players. This is at the cost of a little loss of accuracy for the biggest players (scale $1$) of $ISSIA$ and $TVsoccer$. As we want our final detector to target small players like in $SPD$, we keep $min\_scale=0.1$.\\
     \textit{iii)} As expected, training the 
      student $\mathcal{S}$ on the retrained teacher  annotations $\mathcal{T}^f(\mathcal{X})$
      rather than on the corrected labels $\mathcal{Y}^c$ obtained with the pretrained one 
      improves the results, especially on small players. The intermediate retraining of the teacher permits a first domain adaptation and enables to train the final student $\mathcal{S}$ 
     on more accurate labels. \\
     \textit{iv)} Finally, using context information improves the detection performances at all scales. 

Figures~\ref{fig:teaser} and \ref{fig:tracking_results} show visual results of the student detection, in particular with very small players. Additional examples are shown in the supplementary material.

\subsubsection{Player tracking results}

\paragraph{Fine-tuning of the visual embedding}

We first associate all the detections to tracks using only the spatial consistency IOU criteria. In this particular setting, we find on the validation sequences that the optimal IOU threshold value is $\tau_{IOU}=0.7$. We use this value to create tracks on our unlabeled SoccerNet training sequences. As a post-processing step, we also remove all the tracks with length smaller than $5$ detections since such small tracks are not reliable. 
The generated trajectories are used to fine-tune the object detector network so as to generate inter-player discriminative visual representations. The training is conducted using the triplet loss \eqref{eq:tripletloss} with a margin value $m=2$, $K=5$ 
and $P=10$. 

\paragraph{Inference of the final tracker}

\begin{figure}[h]
\begin{minipage}[b]{0.2\textwidth}
\centering
\includegraphics[width=1.\linewidth]{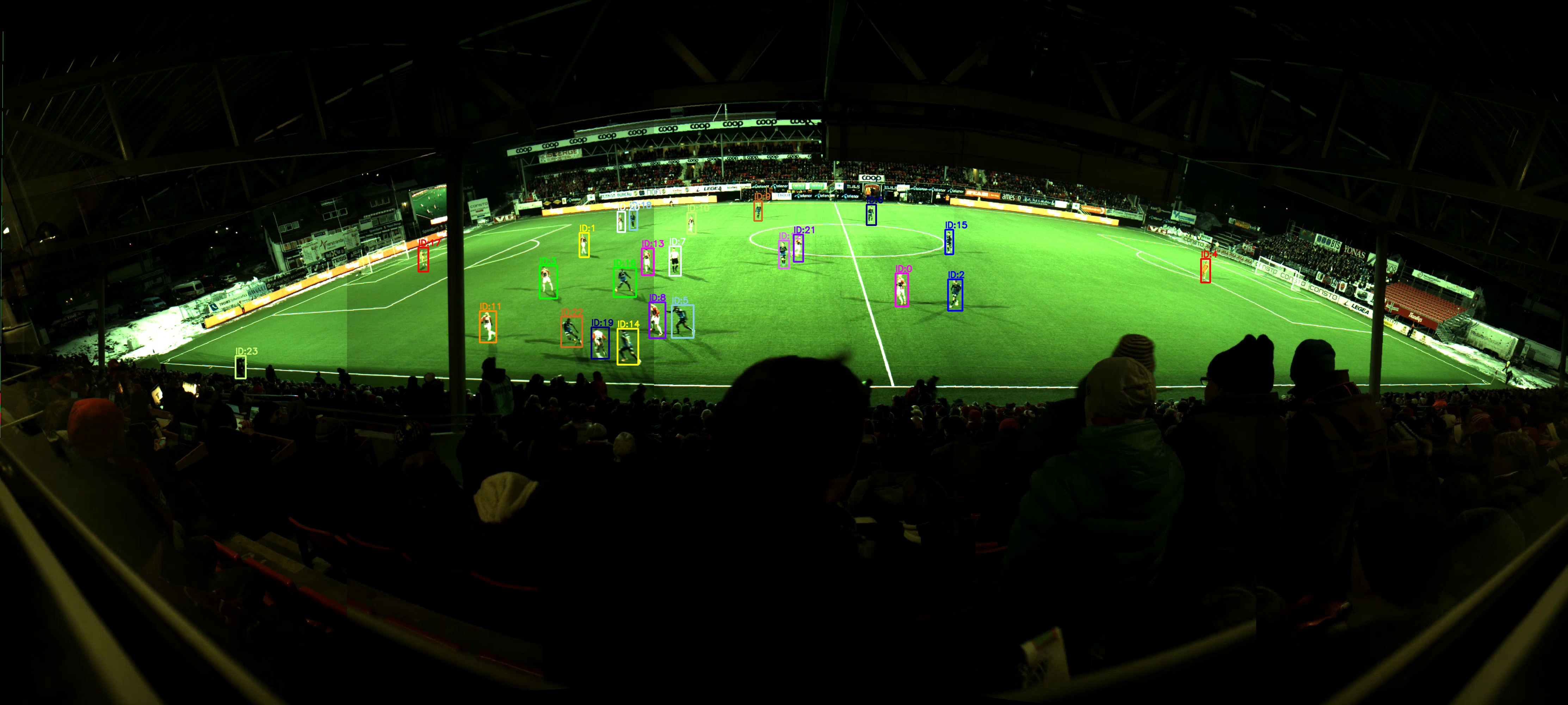}
\end{minipage}
\begin{minipage}[b]{0.2\textwidth}
\centering
\includegraphics[width=1.\linewidth]{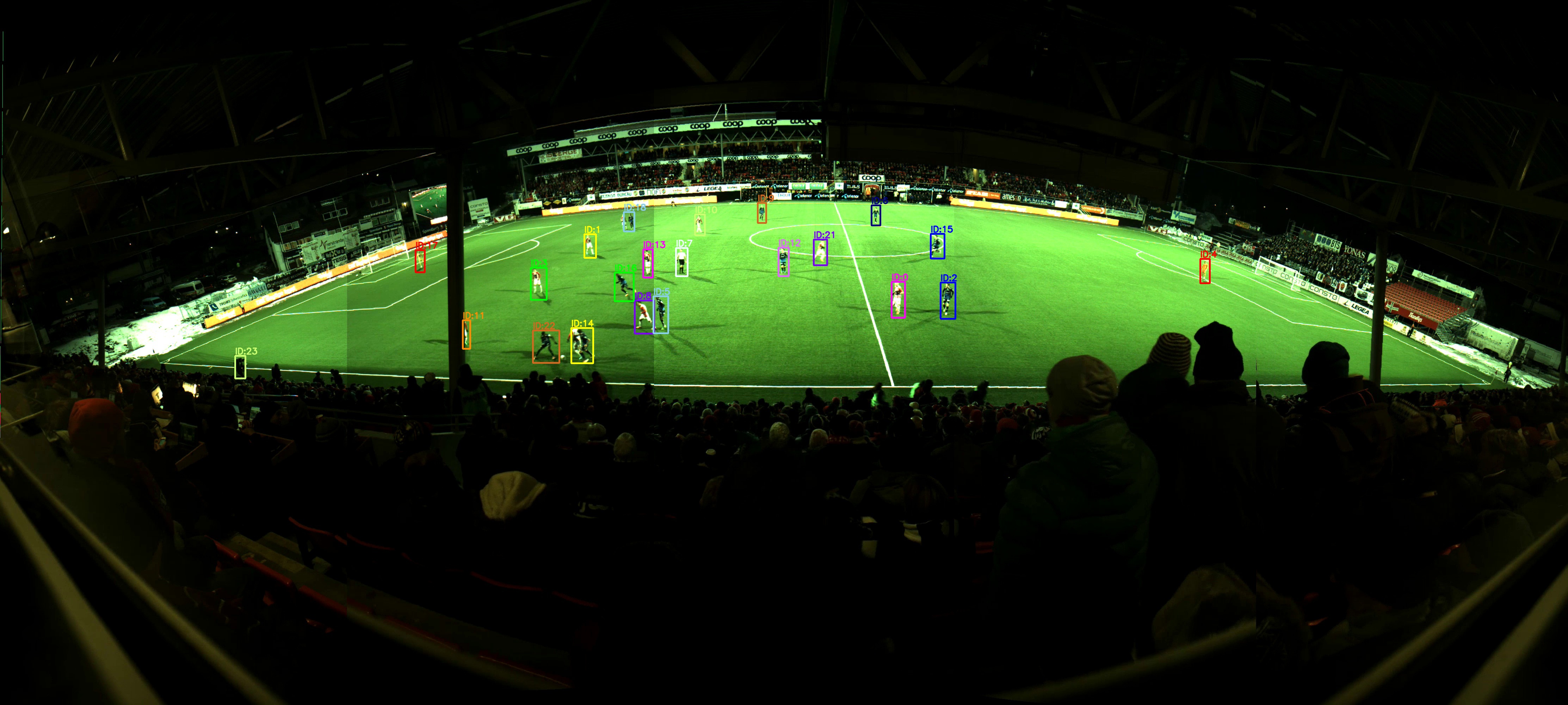}
\end{minipage}
\begin{minipage}[b]{0.2\textwidth}
\centering
\includegraphics[width=\linewidth]{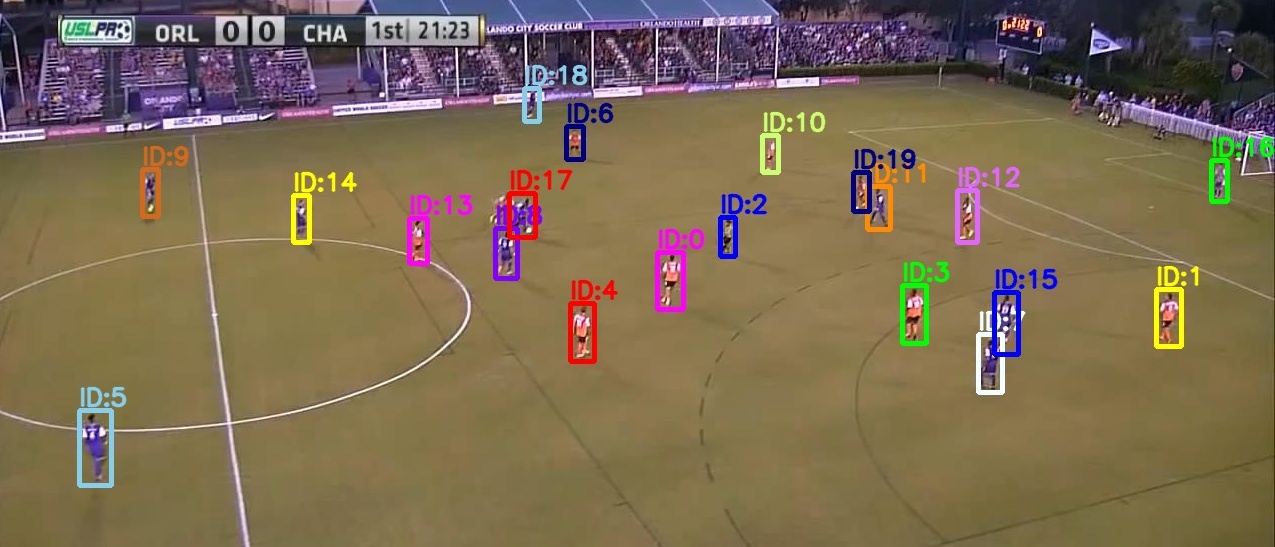}
\end{minipage}
\begin{minipage}[b]{0.2\textwidth}
\centering
\includegraphics[width=\linewidth]{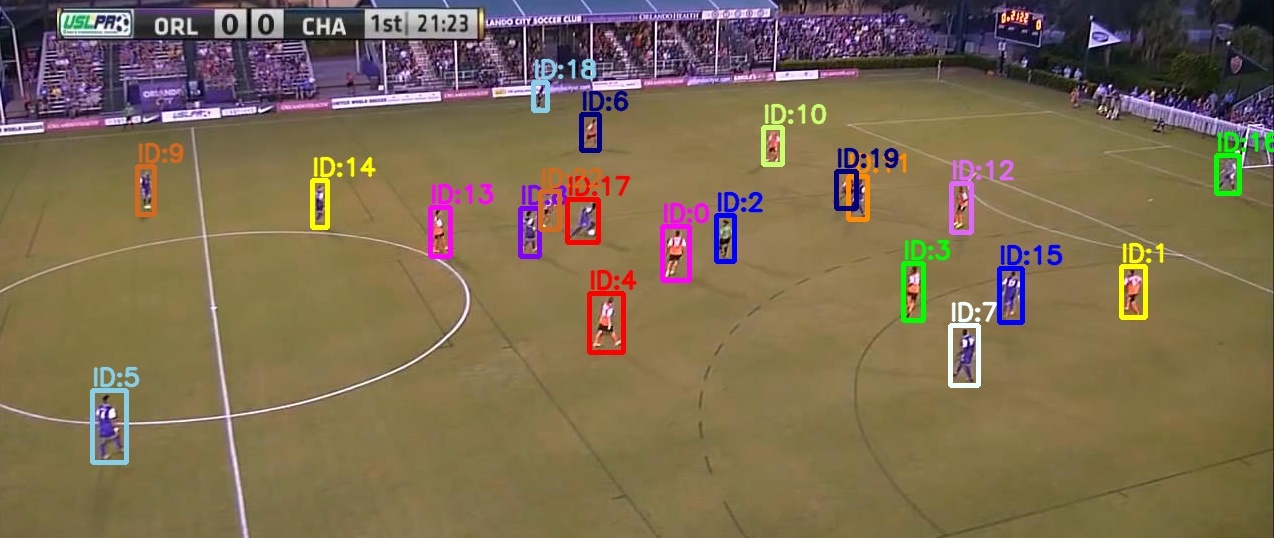}
\end{minipage}
\caption{Visual performances of the proposed tracker on two challenging sequences 
(1 second apart frames).}
\label{fig:tracking_results}
\end{figure}

We report the final results of the tracker in  Table~\ref{tab:result_tracking}. In order to simulate the performances with smaller players, we evaluate the tracking performances at different fixed down-scaling factors. 
For a deeper analysis, we provide three blocks of results. The first one ("No triplet loss FT") displays the tracking performances before triplet loss fine-tuning of the network, and the second one ("Final Tracker") after this retraining step. Notice that it gives a small improvement in MOTA possibly confirming that the performance of the tracker mostly relies on the accuracy of the player bounding box positions. To confirm it, the third block, denoted as "Oracle tracker", shows the performance of the proposed tracking framework applied to ground-truth player positions.  We observe that with ideal bounding boxes, our tracking method  follows all the players almost perfectly. The performance gap between the final tracker and the oracle one is mainly explained by the limits of our detection method in occlusion situations. We illustrate in Figure~\ref{fig:tracking_results} the results of our tracker on two sequences with small players. Additional examples are shown in the supplementary material, including failure cases and a discussion.
\subsection{Comparison with other methods}
\subsubsection{Soccer player detection methods}
Our whole framework can be easily adapted for any given object detection model. The improvements showed in Section \ref{sec:ablation} between the pretrained model and our proposed 
final one prove the efficiency of the proposed method in the case of FPN-FasterRCNN. 
Due to unavailability of data or code, 
it is hard to compare our method with most of the detection and tracking frameworks specialized in 
soccer. However, we can still compare with \cite{FootAndBall} and \cite{cascaded_player} 
using the results given in their respective papers.
Both methods proposed CNN architectures trained in the very same dataset on which they perform evaluation. On the opposite, our method is robust to evaluation on various game conditions. \cite{cascaded_player} does not use standard object detection metrics. We can still compare the precision of their model : on the $SPD$ dataset they announce a precision value of $0.962$ while our model achieves a final precision of $0.97$. 
Table~\ref{tab:comparison_detection} shows that we also improve by large margin the AP performance of \cite{FootAndBall}. 

\subsubsection{Multi-object tracking methods}

We compare in Table \ref{tab:comparison_tracking} our tracker with the state-of-the art Multi-object tracking (MOT)  methods \cite{wang2019towards}  and \cite{bergmann2019tracking}.
Like us, \cite{wang2019towards} uses a triplet loss for learning appearance embedding and a network architecture based on Feature Pyramid Networks. \cite{bergmann2019tracking} is also based on FasterRCNN and associates the detector bounding boxes using spatial information and visual embeddings. As we are interested in the tracking of small players, we report in the first two rows of results of Table~\ref{tab:comparison_tracking} the results of their whole frameworks on the $ISSIA$ dataset with $\times0.5$ down-scaled images. 
The third column "FPS" indicates the computational time for each method in frame per second. Our  proposal outperforms by very large margin these methods. However, even if our method is not trained on any ground-truth data it has been adapted to small soccer player detection. For a fairer comparison, we also compute the tracking results of \cite{bergmann2019tracking} but from the detections given by our detector $\mathcal{S}$. These results, shown in the third row of Table~\ref{tab:comparison_tracking}, indicated by "tracking\_wbw with $\mathcal{S}$", display a clear boost on their  performance. Our proposed tracking method and \cite{bergmann2019tracking} obtain really close results on the detections given by our $\mathcal{S}$ detector, corroborating the benefits of our detector. Our approach is however simpler, as we do not use any additional re-id network and is computationally two times faster. Moreover, our method is unsupervised while the tracking extension (re-id module) in \cite{bergmann2019tracking} has  been trained with ground truth tracking data. Finally, let us remark that the combination of our detector with the architecture of \cite{wang2019towards} is not possible as their model integrates detection and tracking in a joint estimation. 

\begin{table}
  \caption{MOTA and IDF1 performance of our final MOT on ISSIA dataset, for different 
  down-scaling factors. 
  Extra blocks of results provide a further analysis (details in the text).}
  \label{tab:result_tracking}
    \begin{tabular}{c|c|c|c}
    Method & Scaling factor & MOTA & IDF1\\
    \midrule
    No triplet loss FT & 1. & 87.3 & 74.6 \\
                        & 0.75 & 92.3 & 82.8\\
                        & 0.5 & 93.1 &  81.7 \\
    \midrule                  
    Final tracker    & 1. & 87.7 & 76.6 \\
                        & 0.75 &  93.4 & 82.3 \\
                        & 0.5 & 92.9 & 80.9 \\
    \midrule       
    Oracle tracker    & 1 & 99.9 & 99.3 \\
                    & 0.75 & 99.7 & 95.0 \\
                    & 0.5 & 100. & 96.8 \\
    \bottomrule
\end{tabular}
\end{table}

\begin{table}
  \caption{AP comparison of the proposed detector 
  with the soccer player detection method \cite{FootAndBall}.}
  \label{tab:comparison_detection}
\begin{tabular}{c|c|c|c}
    Dataset & $ISSIA$ & $SPD$ \\
    \midrule
    FootAndBall \cite{FootAndBall} & 92.1 & 88.5 \\
    Ours & {\bf 97.3} & {\bf 98.0} \\
   \bottomrule
\end{tabular}
\end{table}

\begin{table}
  \caption{IDF1, MOTA and running time (frames per second, FPS)  comparison of the proposed MOT with the 
  methods \cite{wang2019towards} and \cite{bergmann2019tracking}. Best results in {\bf bold} and second best \underline{underlined}.}
  \label{tab:comparison_tracking}
    \begin{tabular}{c|c|c|c}
    
    Method & MOTA & IDF1  & FPS \\
    \midrule
    towards\_rt\_mot \cite{wang2019towards} & 46.4 & 24.7 &  {\bf 13.1} \\
    tracking\_wbw \cite{bergmann2019tracking} & 23.5 & 27.8 &  3.4 \\
    tracking\_wbw with our detections & \underline{ 91.0} & {\bf 85.7} &  3.4 \\
    Ours & {\bf 92.9} & \underline{80.9} &  \underline{6.5} \\
   \bottomrule
\end{tabular}
\end{table}

\section{Conclusion} \label{sect:conclusions}
This paper presents a deep-learning-based soccer player detection and tracking method robust to small players and assorted conditions (such as different stadiums, teams, viewpoints and illumination settings). As a key component, our method is trained in an unsupervised way, so no manual annotations are  needed. We have performed a thorough quantitative analysis of the different components of the method while highlighting the key ones. The comparison with state-of-the-art methods, both in detection and tracking, shows that our method offers competitive results, especially in the detection stage. 
As future work, we plan to improve the tracking results in 
challenging cases such as long-time occlusions.

\begin{acks}
SH acknowledges support by ENS Paris-Saclay. CB and GH
acknowledge support by MICINN/FEDER UE project, ref. PGC2018-098625-B-I00, H2020-MSCA-RISE-2017 project, ref. 777826 NoMADS, and RED2018-102511-T. 
We thank Patricia Vitoria for her comments that greatly improved the manuscript. 
\end{acks}

\bibliographystyle{ACM-Reference-Format}
\bibliography{main}

\clearpage
\newpage
\appendix
\subtitle{Supplementary Material}

\section{Training data}

\begin{figure}[ht]
\begin{minipage}[b]{0.2\textwidth}
\centering
\includegraphics[width=\linewidth]{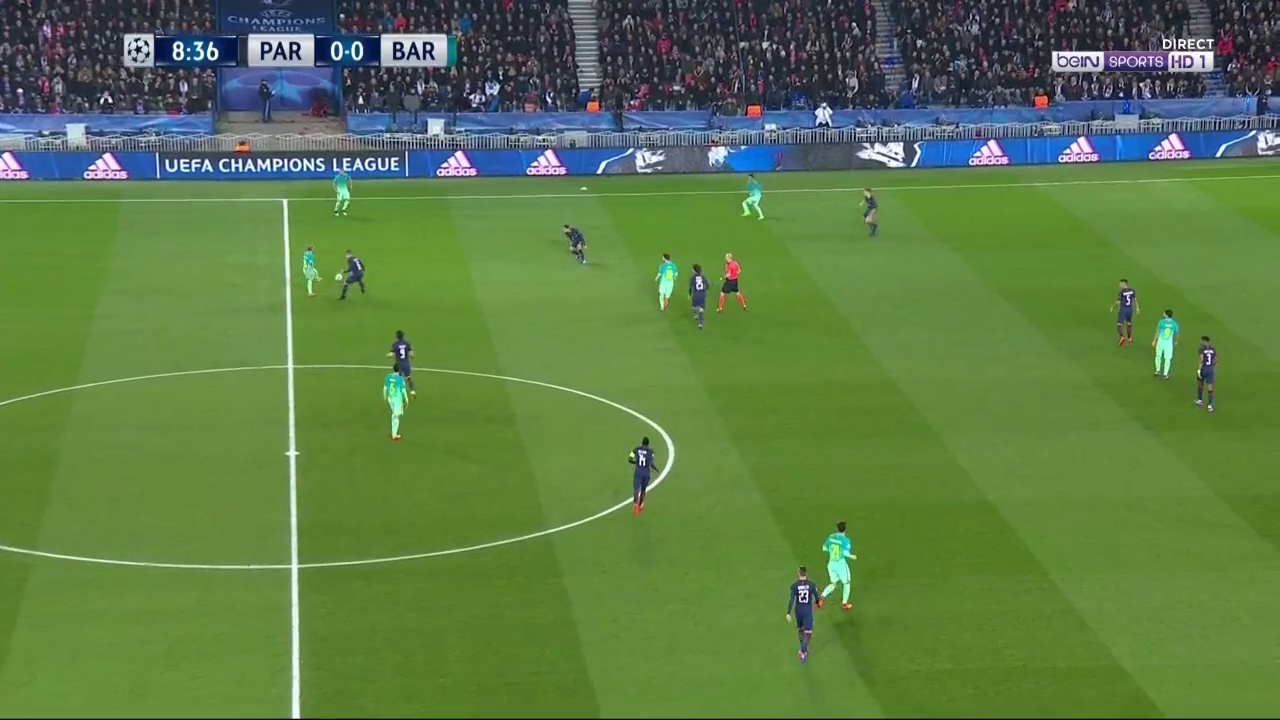}
\end{minipage}
\begin{minipage}[b]{0.2\textwidth}
\centering
\includegraphics[width=\linewidth]{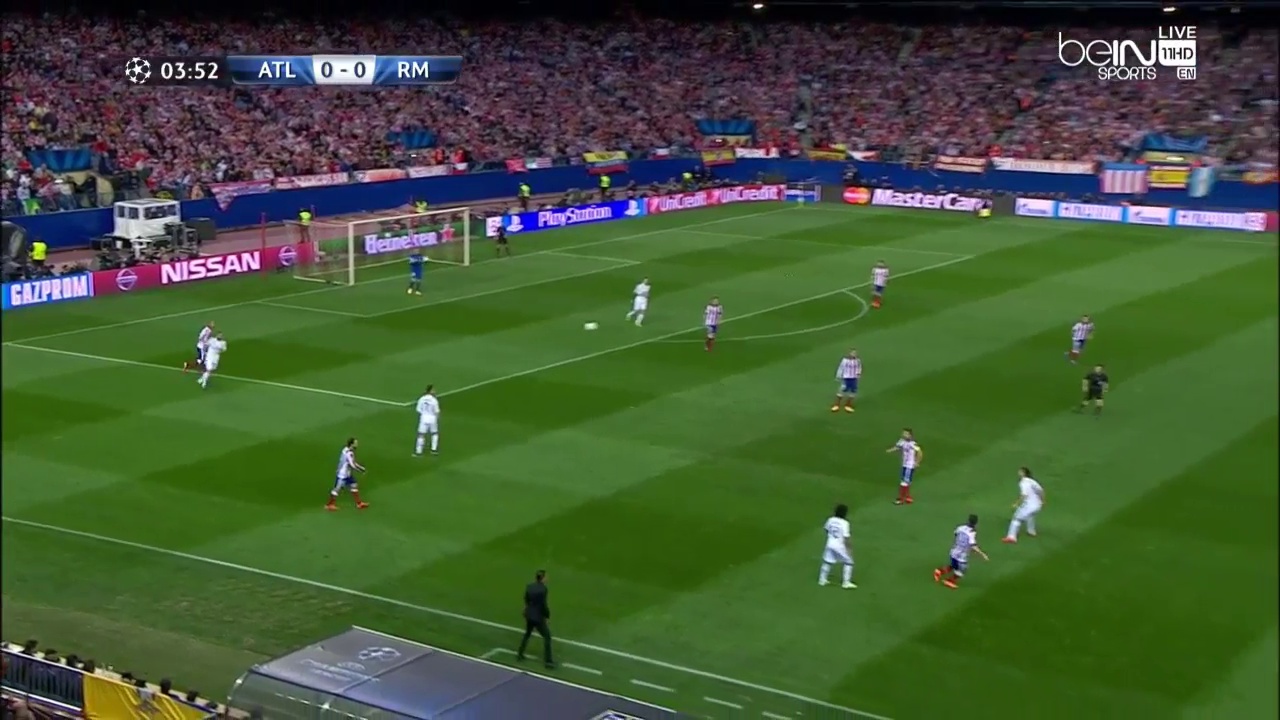}
\end{minipage}
\caption{Examples of SoccerNet training images. }
\label{SoccerNet_examples}
\end{figure}

Our training images are unlabeled TV broadcast soccer videos taken from a subset of the SoccerNet dataset, where we extract only wide views. Figure \ref{SoccerNet_examples} and the left column of Figure \ref{field_detection_im} show some examples of these training images.

\section{Details of the method}

In this section, we detail the algorithms used to remove false positives and add false negatives detections to the teacher $\mathcal{T}^{init}$ annotations. These methods have to be as accurate as possible and can be, to a certain extent, computationally expensive as they are not used for the final inference of the student $\mathcal{S}$.

\subsection{Field detection}

As a first step for detecting the field we find a mask that contains the potential pixels of the field. Then, it is successively refined. As observed in Figure~\ref{SoccerNet_examples}, coaches are often present in the green bottom part of the field, and those should not be included in the mask.
The field mask is computed thanks to following successive processing steps :  

\begin{itemize}
    \item Application of a green filter to create a first binary image. We use as upper and lower color threshold the values (15,50,50) and (70,255,255).
    \item Selection of the contour with the biggest connected component.
    \item Approximation of the contour by a polygon.
    \item Detection of the bottom line of the field and removal of the mask pixels below it.
\end{itemize}

\begin{figure}[ht]
\begin{minipage}[b]{0.2\textwidth}
\centering
\includegraphics[width=\linewidth]{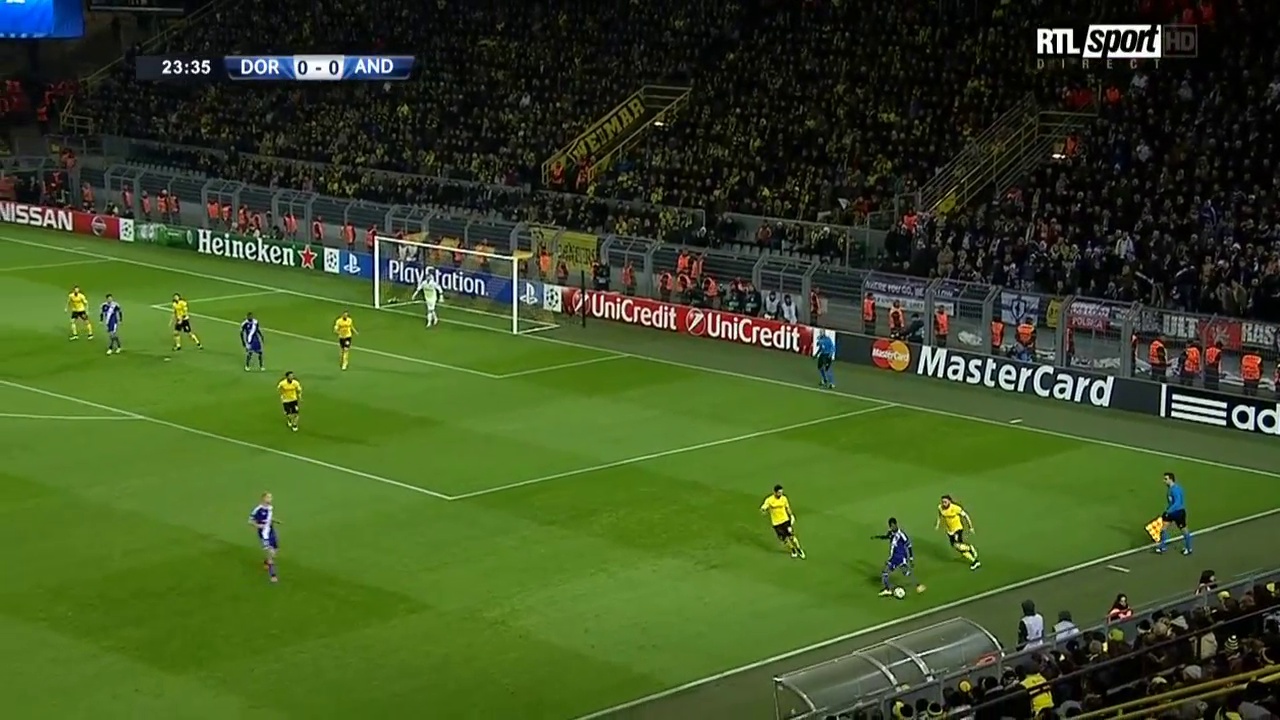}
\end{minipage}
\begin{minipage}[b]{0.2\textwidth}
\centering
\includegraphics[width=\linewidth]{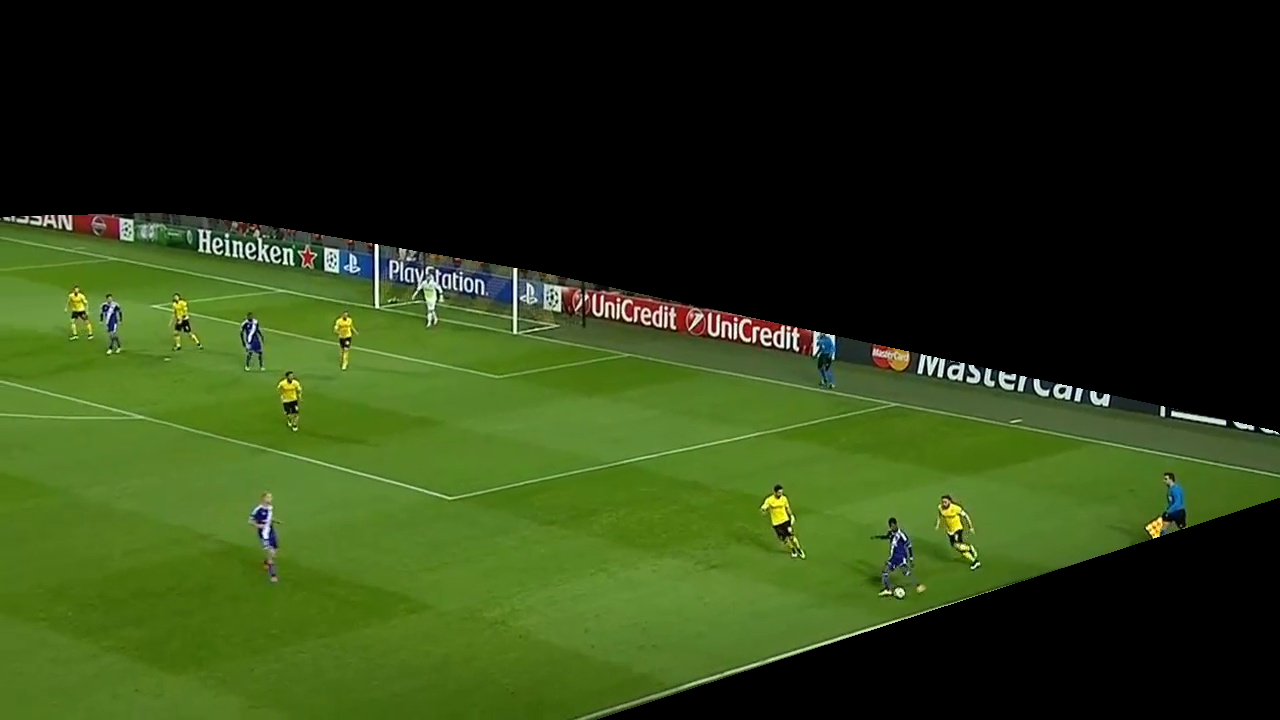}
\end{minipage}
\begin{minipage}[b]{0.2\textwidth}
\centering
\includegraphics[width=\linewidth]{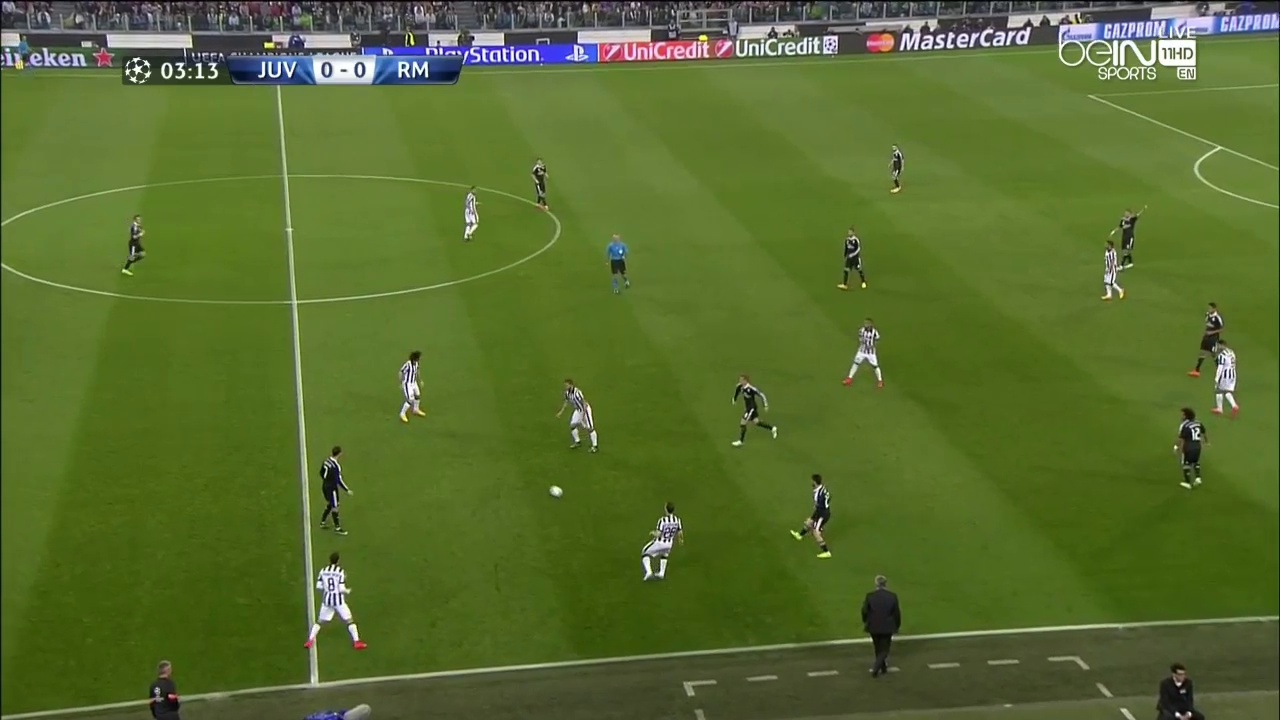}
\end{minipage}
\begin{minipage}[b]{0.2\textwidth}
\centering
\includegraphics[width=\linewidth]{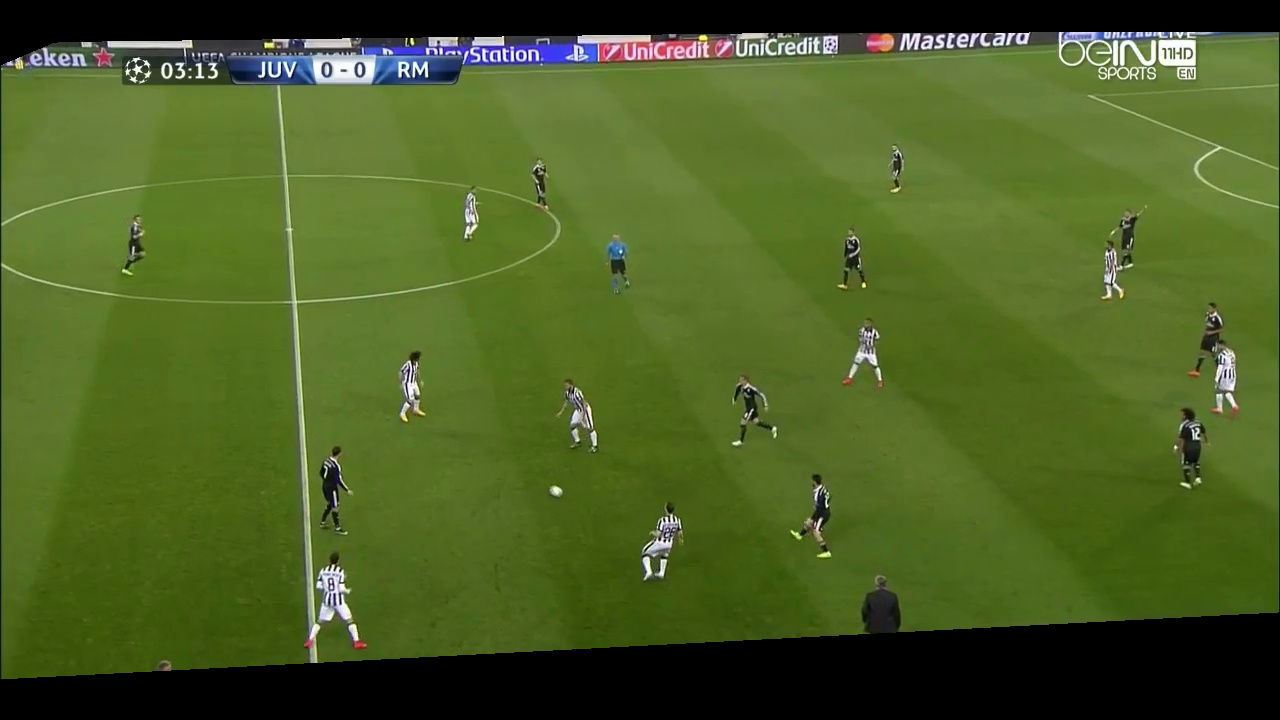}
\end{minipage}
\begin{minipage}[b]{0.2\textwidth}
\centering
\includegraphics[width=\linewidth]{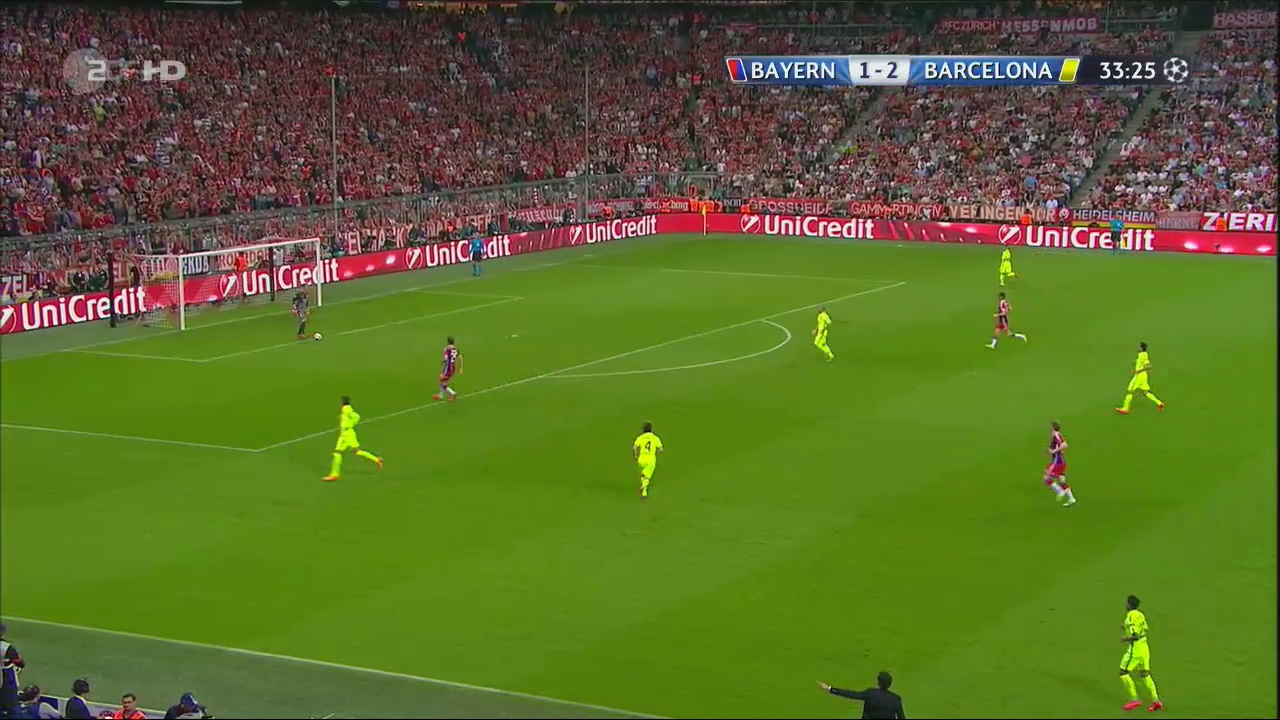}
\end{minipage}
\begin{minipage}[b]{0.2\textwidth}
\centering
\includegraphics[width=\linewidth]{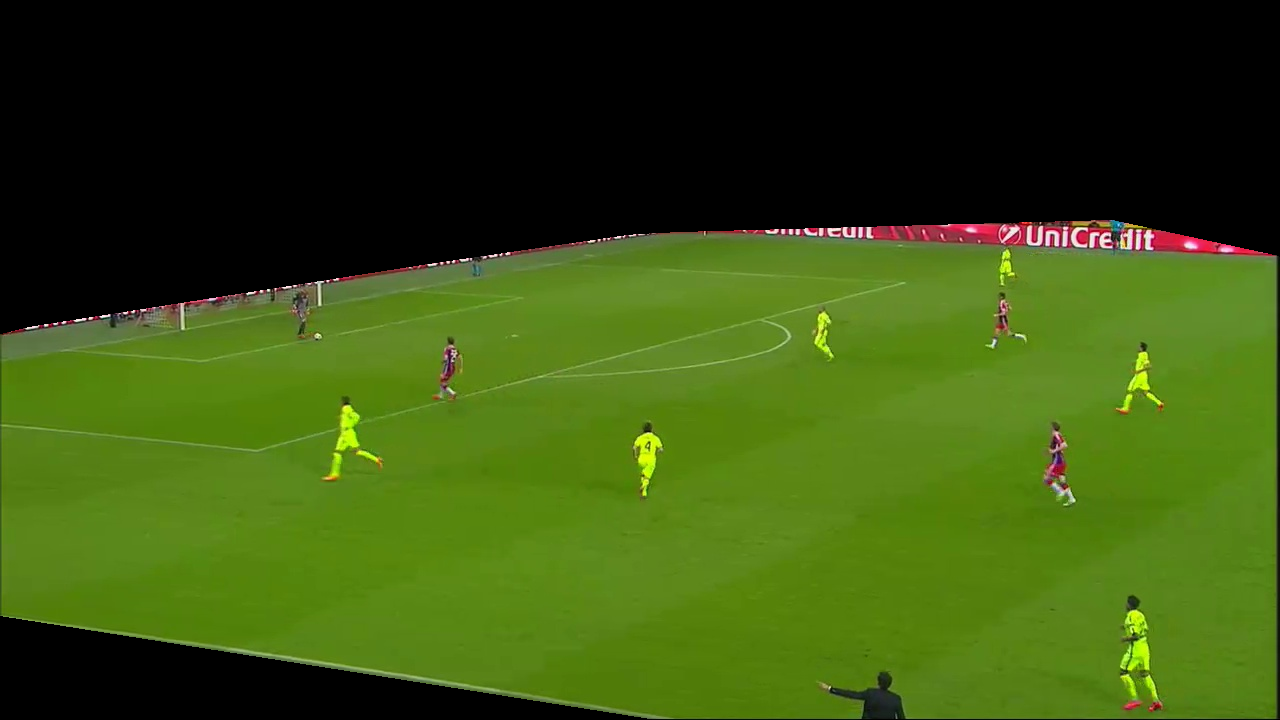}
\end{minipage}
\caption{Three results obtained by the proposed field detection approach. Left: original image. Right: Output of our field detector.}
\label{field_detection_im}
\end{figure}

To detect the bottom line we use the \cite{line_pix2pix} pix2pix model to get a binary image approximating the lines of the field. We then  run a Hough line transform on this binary image and use thresholds on the length, angle and position of the detected lines.  Some examples of the final estimated field mask are shown in Figure~\ref{field_detection_im}. 

\subsection{Add missed players to the training data}
\label{sec:blob}

We detail here the blob detection strategy used to add players to the training data. Given an input image $u$, we work on the pixels belonging to the previously extracted field mask. Note that the teacher $\mathcal{T}^{init}$ has already detected some of the players $\mathcal{T}^{init}(u)$. 

We first compute blobs through green filtering and contour selection. We select the contours for which the enclosing bounding box verifies~: $i)$  it has a similar area as the bounding boxes of the players already detected by the teacher $\mathcal{T}^{init}$ and ii) it does not exist any detection in $\mathcal{T}^{init}(u)$ with intersection-over-union $IOU<0.2$.

We then run a human pose estimation algorithm \cite{chen2018cascaded} on the corresponding enlarged extracted regions. If the confidence of the pose detection  is higher than $0.8$, it is considered as a player and it is added to the annotations. We have observed in our experiments that the pretrained version of the human pose estimator \cite{chen2018cascaded} performs better on small humans than the multi-person  FPN-FasterRCNN model, $\mathcal{T}^{init}$. 

\begin{figure}[ht]
\begin{minipage}[b]{0.2\textwidth}
\centering
\includegraphics[width=\linewidth]{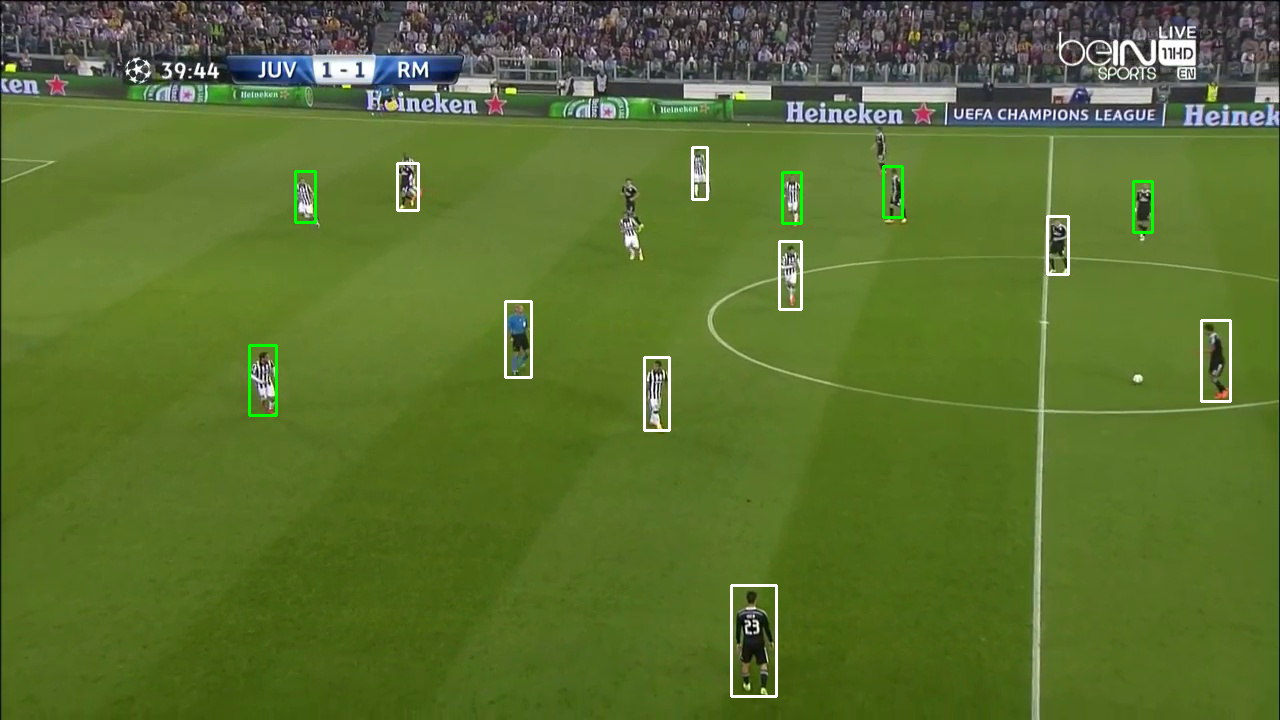}
\end{minipage}
\begin{minipage}[b]{0.2\textwidth}
\centering
\includegraphics[width=\linewidth]{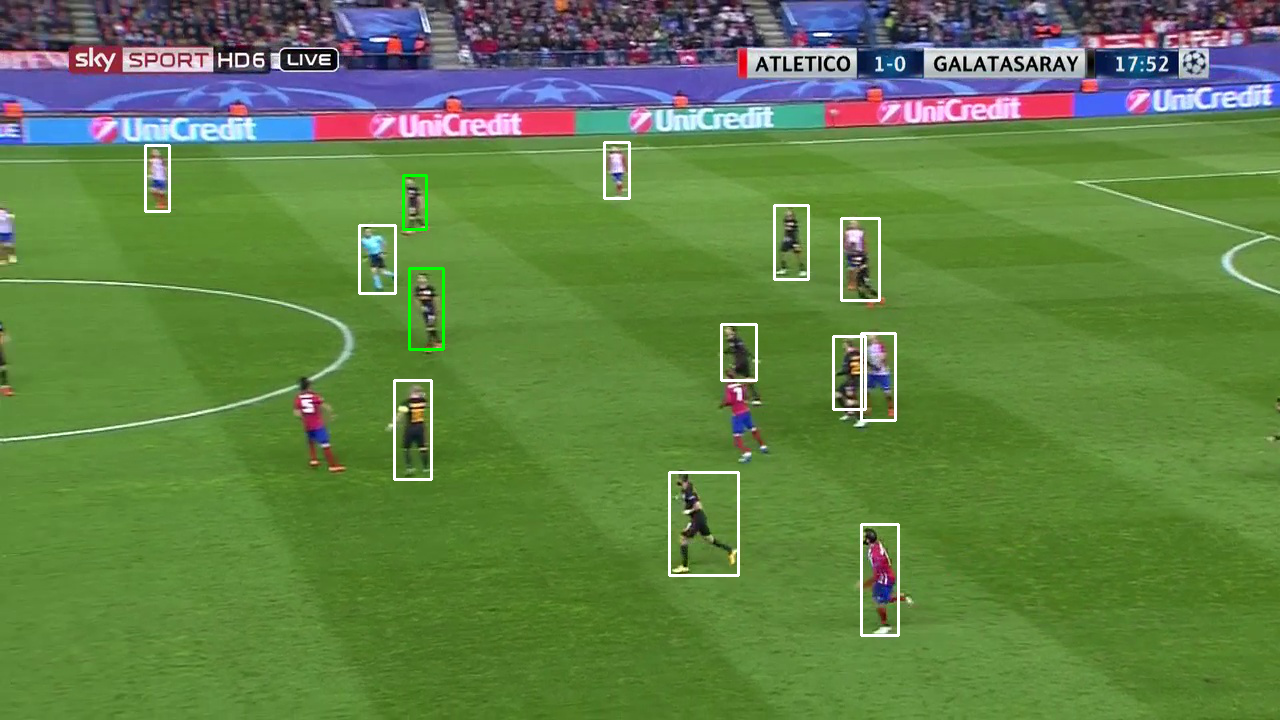}
\end{minipage}
\begin{minipage}[b]{0.2\textwidth}
\centering
\includegraphics[width=\linewidth]{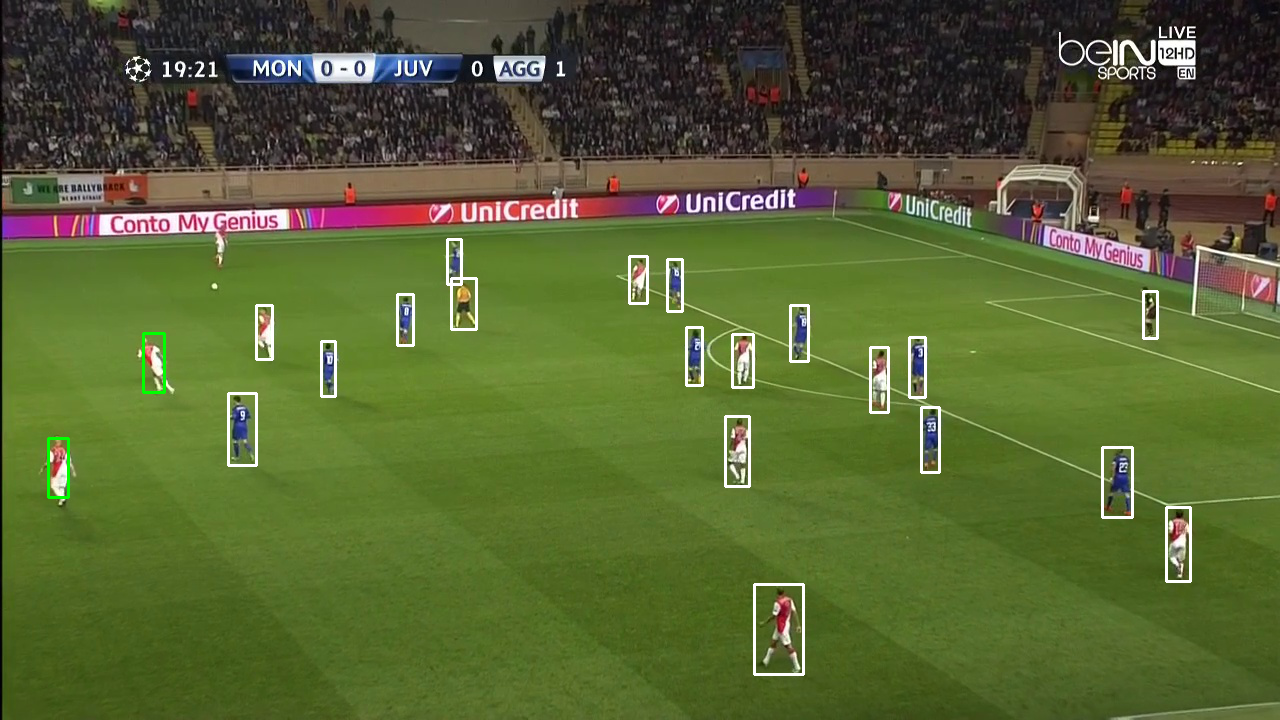}
\end{minipage}
\begin{minipage}[b]{0.2\textwidth}
\centering
\includegraphics[width=\linewidth]{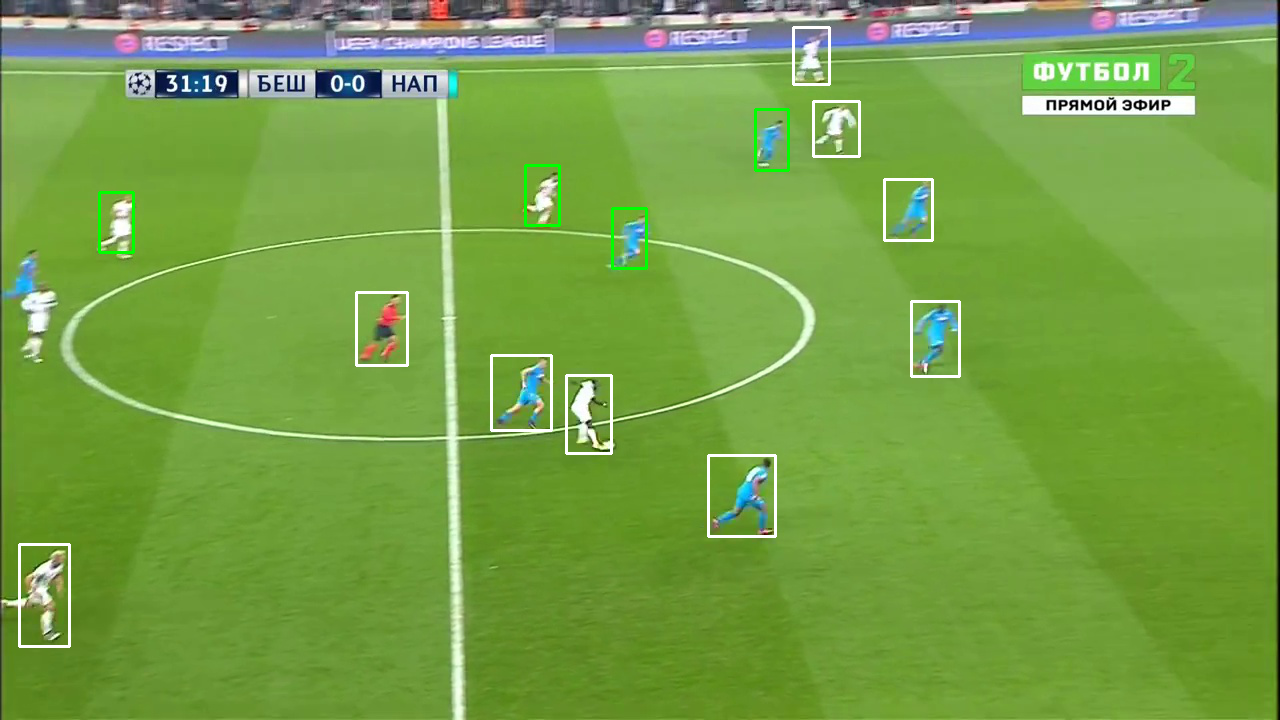}
\end{minipage}
\caption{Examples of training images with the annotations of $\mathcal{Y}^c$. In white the annotations of the pretrained teacher $\mathcal{T}^{init}$ and in green the added players detected with the strategy detailed in Section \ref{sec:blob} to include missing players.}
\label{training_data}
\end{figure}

We display in Figure~\ref{training_data} some examples of the training images with corrected annotations $\mathcal{Y}^c$. 

\section{Implementation details}

We indicate here additional parameter values involved in the proposed framework : \\
i) We train the teacher and student detectors during 20 epochs using the SGD optimizer with an initial learning rate of $10^{-3}$ decayed with a factor $0.1$ at epochs 8, 14 and 17. \\
ii) The fine-tuning with triplet loss is done on $5$ epochs with initial learning rate of $10^{-4}$. \\
iii) For tracking inference we find the optimal parameters to be~: $N_{reID}=10$; $\alpha = 0.03$; $D_{visual\_max}=4$ and $D_{spatial\_max}=(1/16)*image\_width$.

\section{Quantitative results}

In Section 4.5.1 of our paper, Table 5 shows that  the Resnet18 backbone outperforms Resnet50 on the $SPD$ dataset. We made the hypothesis that it is due to the fact that there exists a small difference between the appearance of the players in $SPD$ and SoccerNet training images.  

In Table~\ref{tab:FT} we propose to fine-tune $\mathcal{T}^{f}$ (Resnet50 backbone) and $\mathcal{S}$ (Resnet18 backbone) on $80\%$ of the $SPD$ dataset. We then evaluate on the remaining $20\%$. With this fine-tuning it is now the Resnet50 backbone that outperforms its Resnet18 counterpart. This test confirms that the Resnet50 backbone slightly overfits on the typical TV broadcast player appearances of the training data. Besides providing a faster inference, using a smaller backbone allows the student $\mathcal{S}$ to better generalize. 

\begin{table}
  \caption{AP on $SPD$ of fine-tuned $\mathcal{T}^{f}$ and $\mathcal{S}$ before and after fine-tuning on $SPD$ images.}
  \label{tab:FT}
\begin{tabular}{c|c|c}
    Model &  $\mathcal{T}^{f}$ & $\mathcal{S}$ \\
    \midrule
    Before fine-tuning & 92.0 & 96.0 \\
    After fine-tuning & 99.3 & 98.9 \\
   \bottomrule
\end{tabular}
\end{table}

\section{Visual results}

\subsection{Player detection}

\begin{figure*}[ht]
\begin{subfigure}[b]{\textwidth}
\centering
\includegraphics[width=0.3\linewidth]{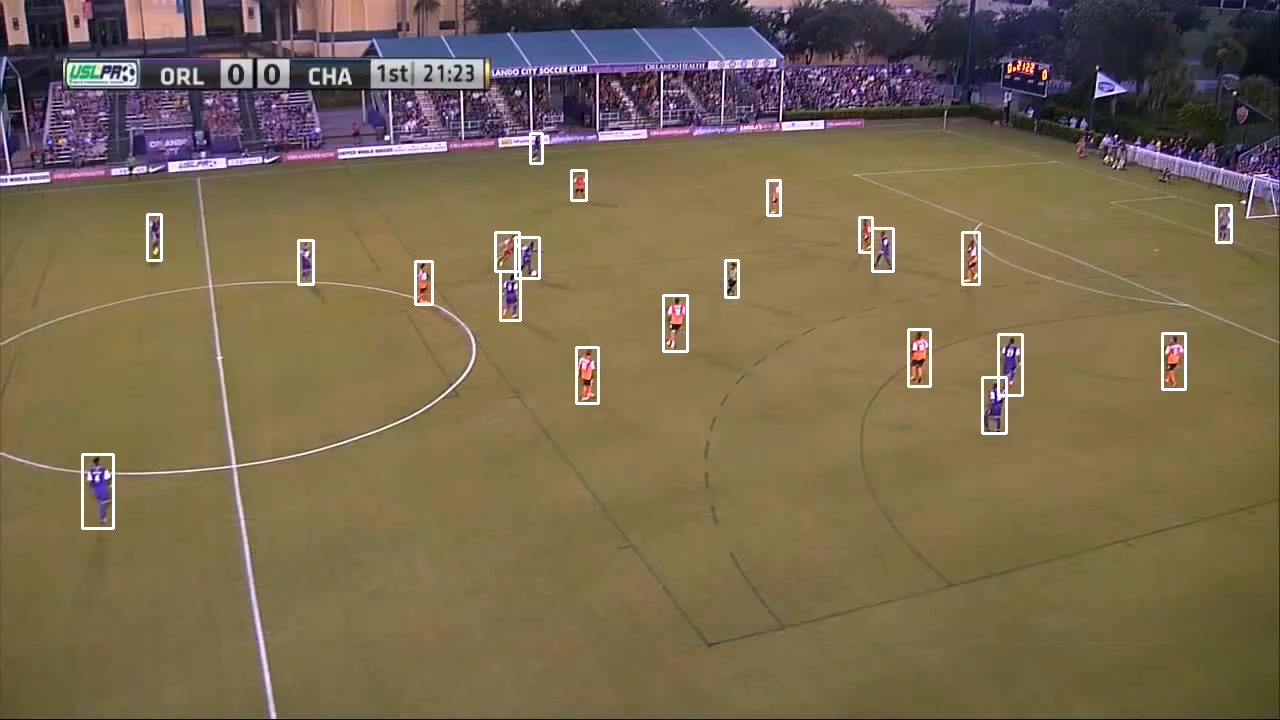}
\includegraphics[width=0.3\linewidth]{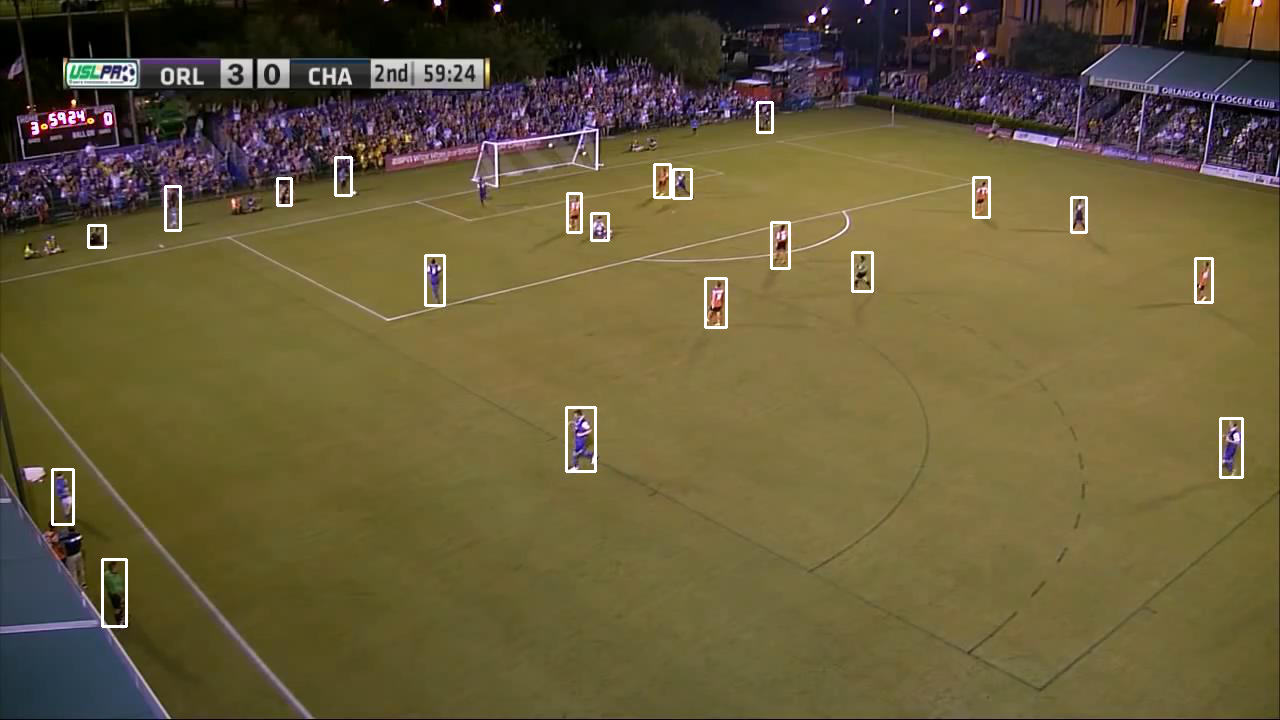}
\includegraphics[width=0.3\linewidth]{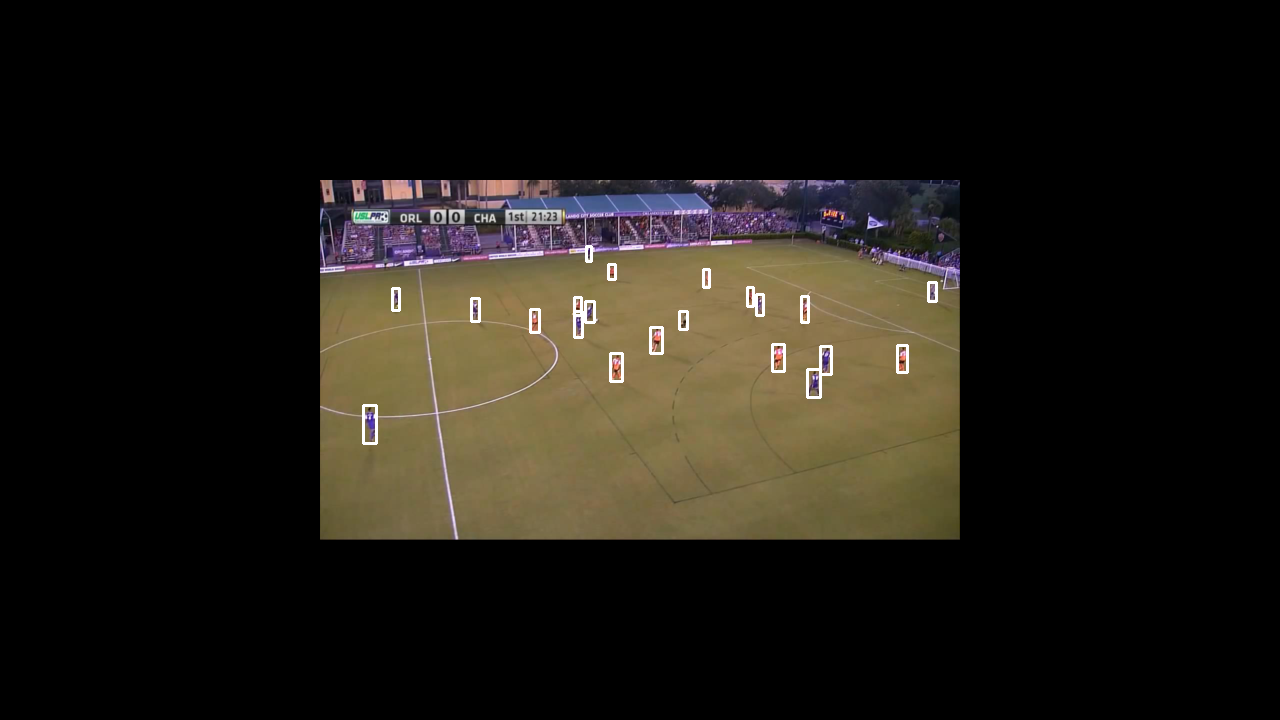}
\caption{$SPD$ dataset}
\end{subfigure}
\begin{subfigure}[b]{\textwidth}
\centering
\includegraphics[width=0.3\linewidth]{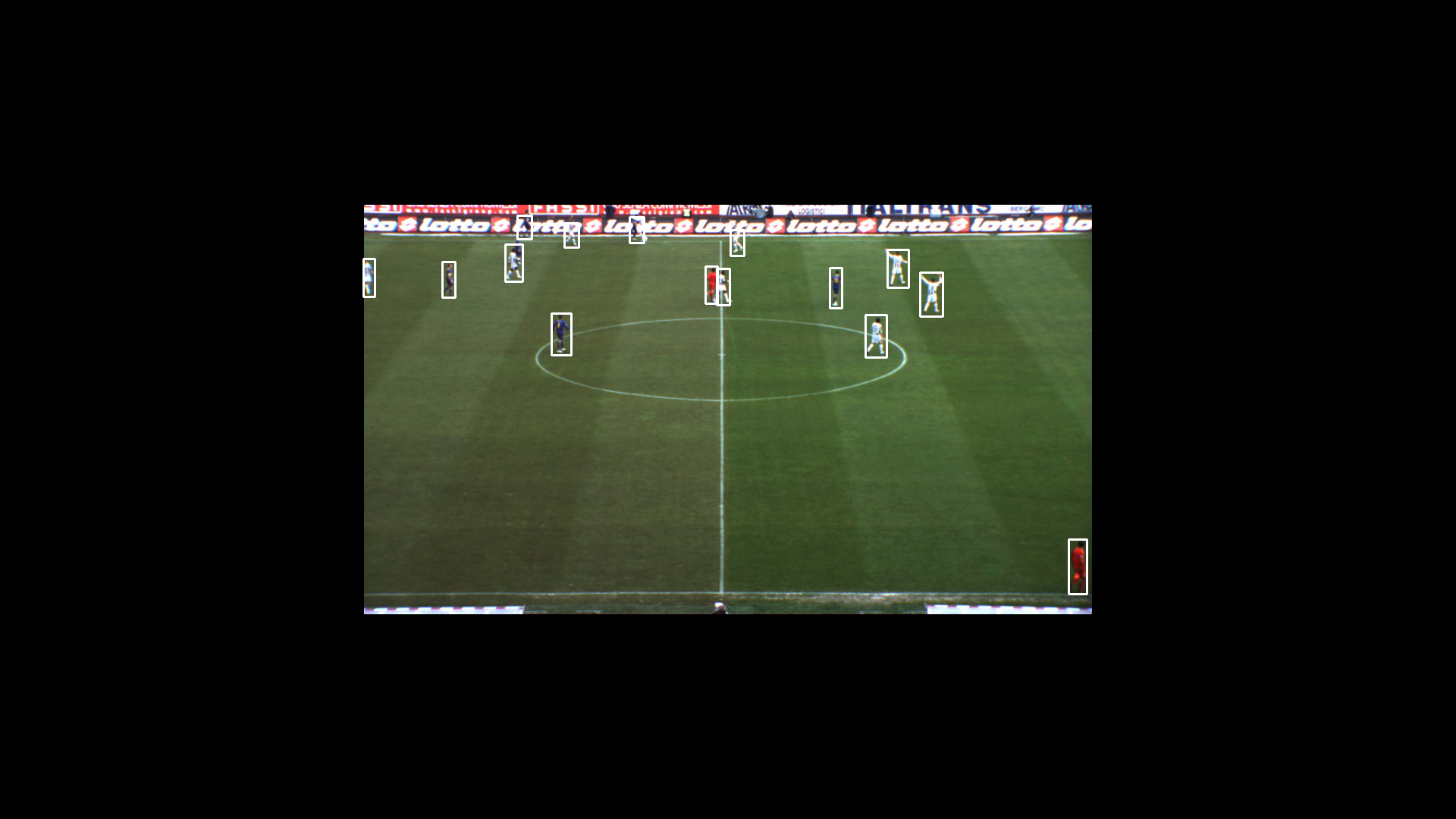}
\includegraphics[width=0.3\linewidth]{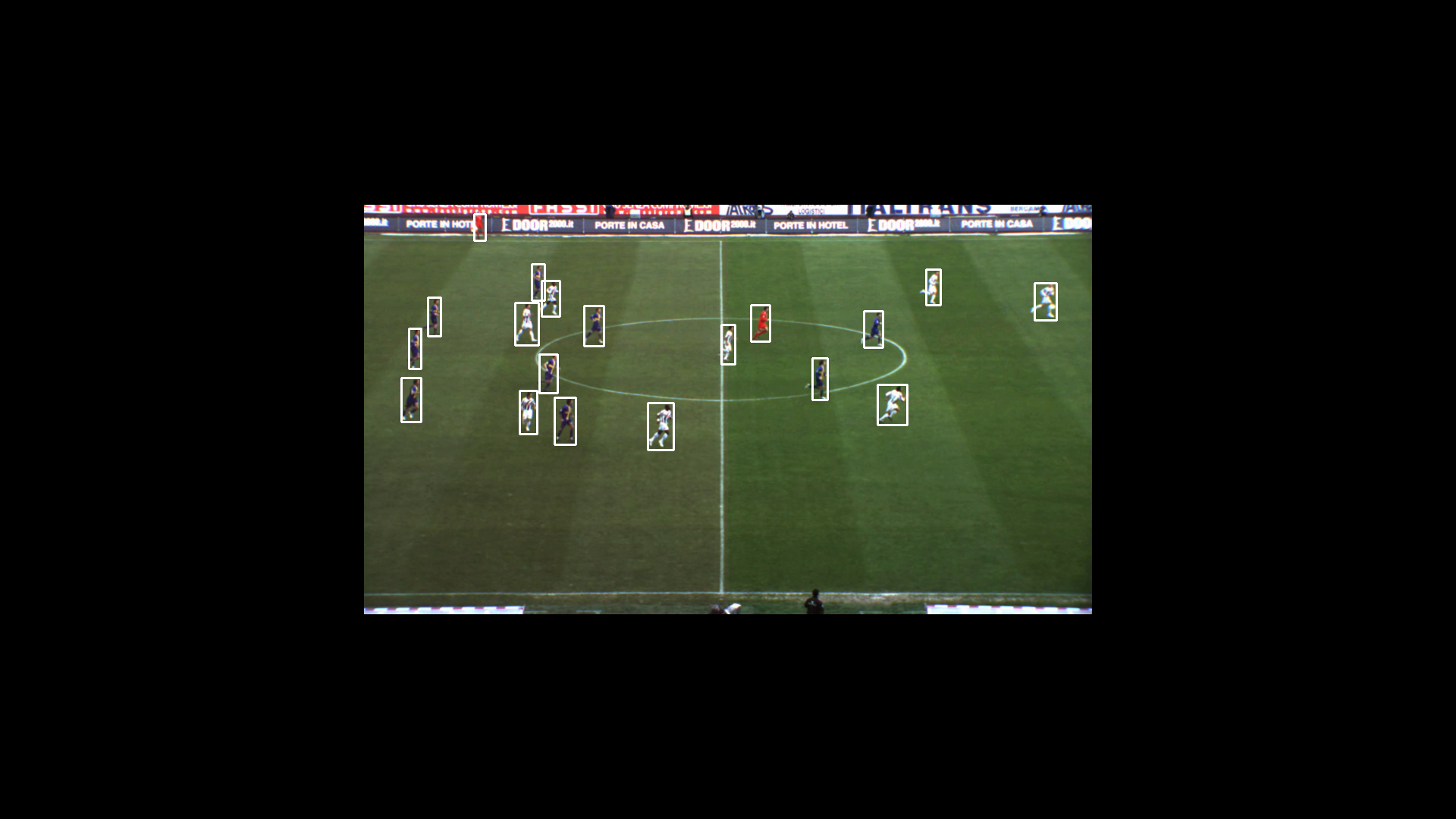}
\includegraphics[width=0.3\linewidth]{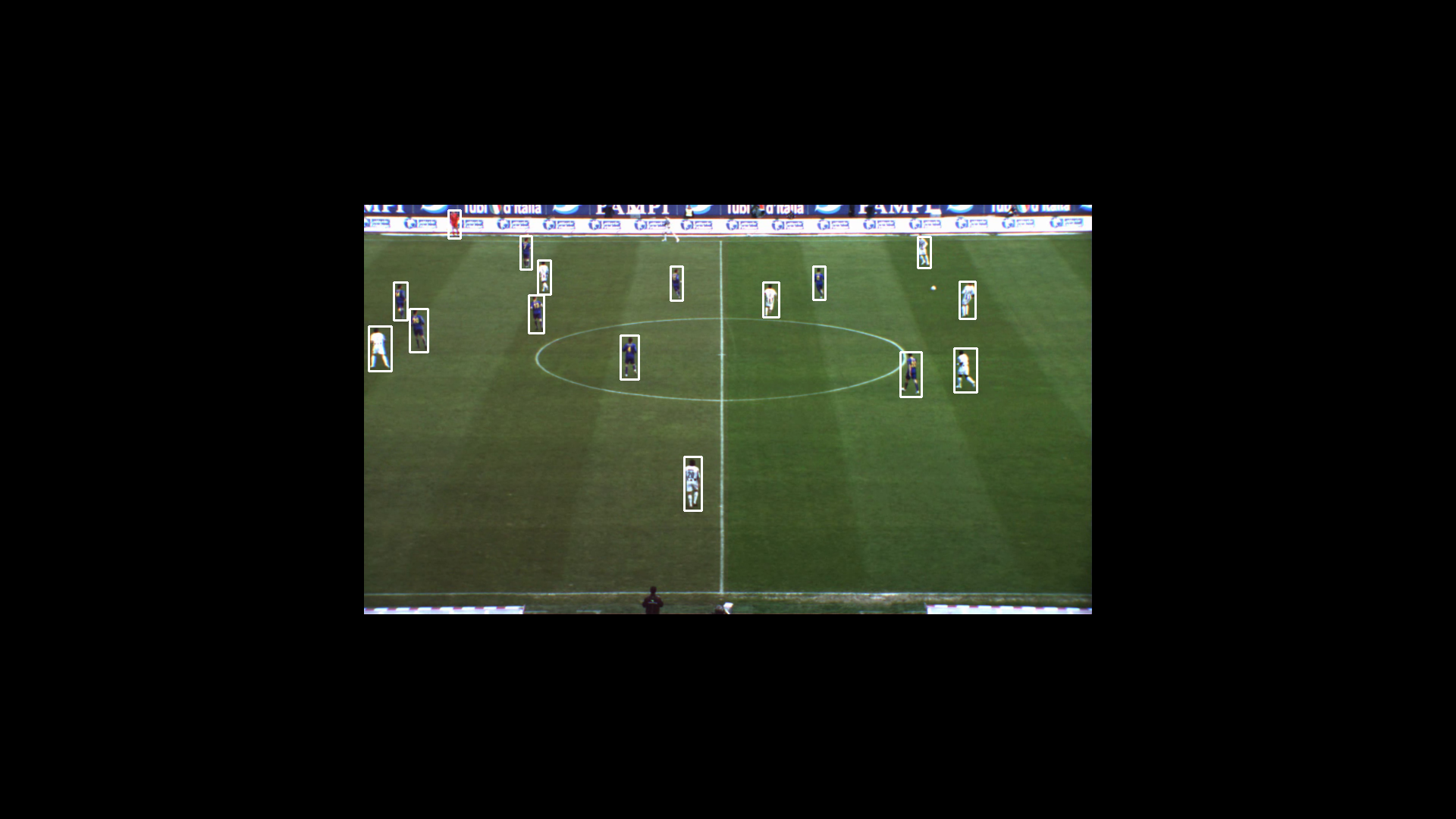}
\caption{$ISSIA$ dataset with down-scaling $x0.5$}
\end{subfigure}
\begin{subfigure}[b]{\textwidth}
\centering
\includegraphics[width=0.3\linewidth]{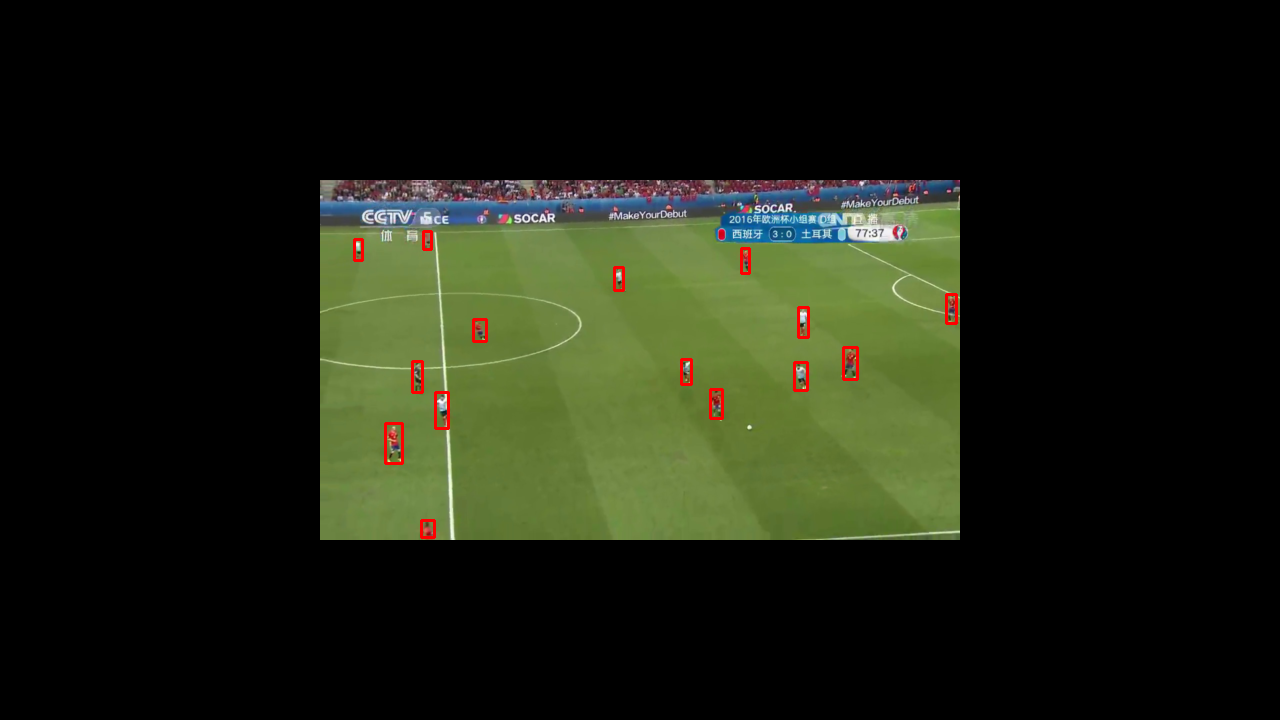}
\includegraphics[width=0.3\linewidth]{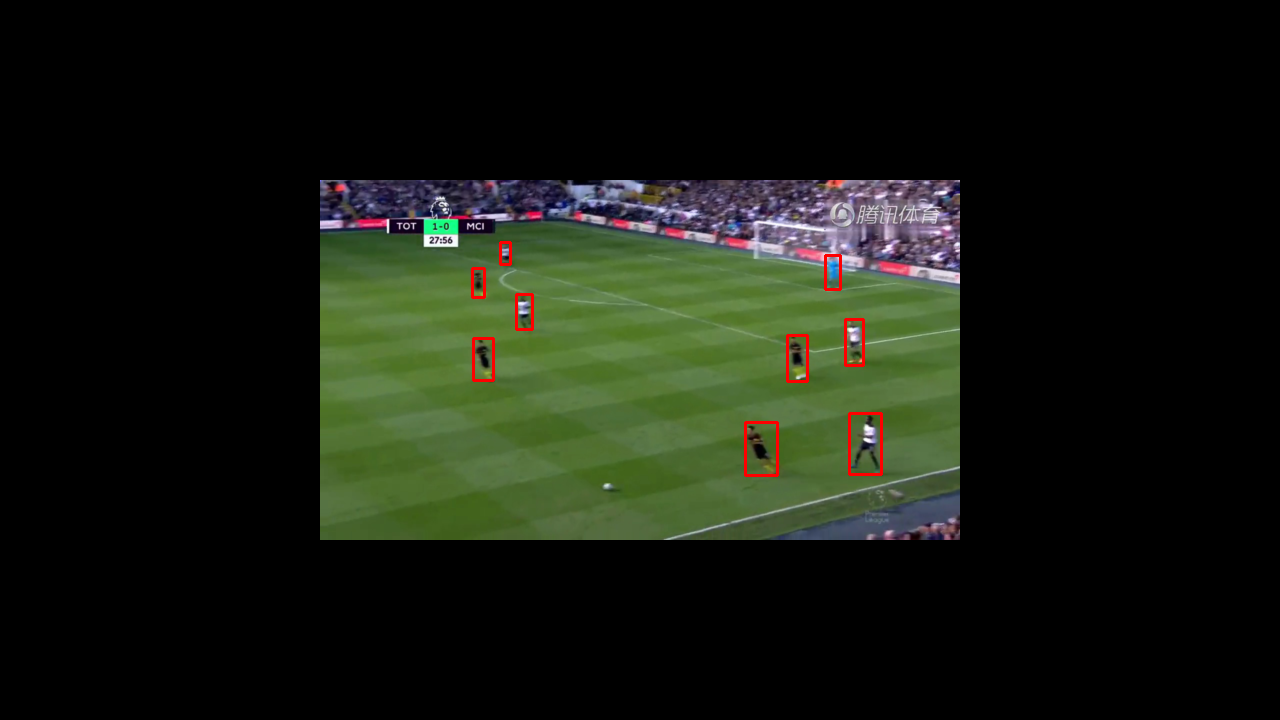}
\includegraphics[width=0.3\linewidth]{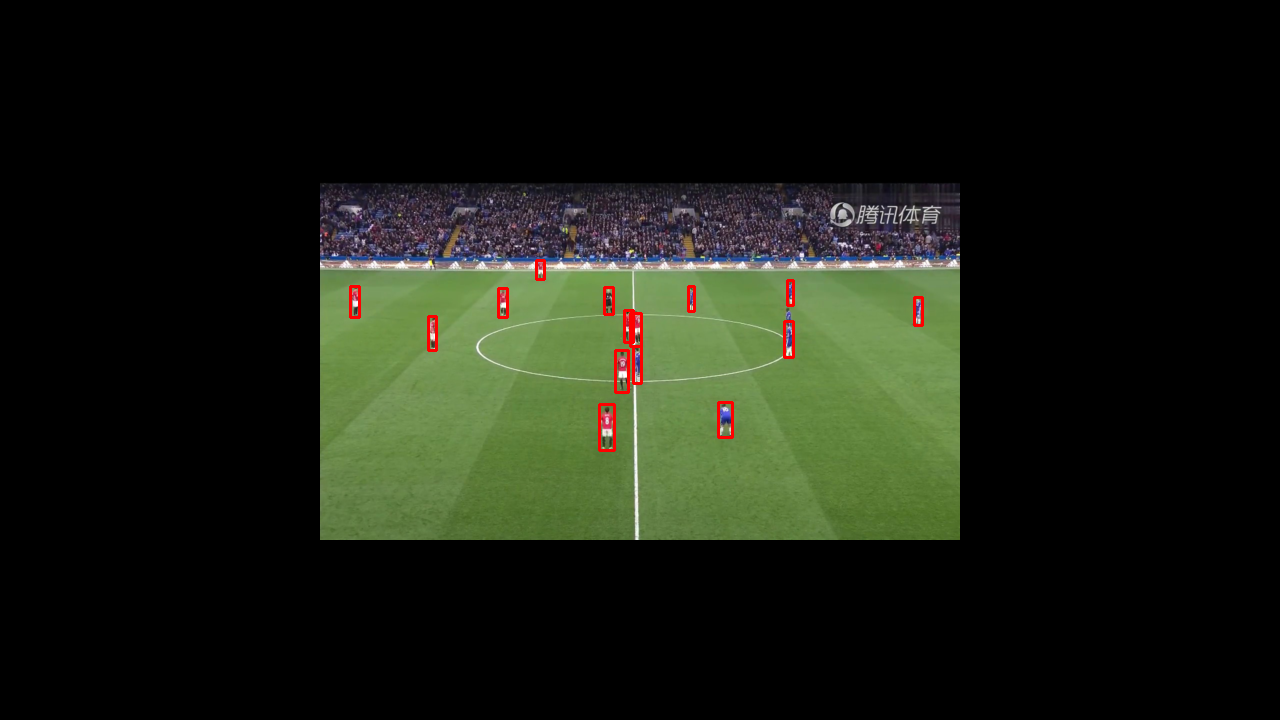}
\caption{$TV\_soccer$ dataset with down-scaling $x0.5$}
\end{subfigure}
\begin{subfigure}[b]{\textwidth}
\centering
\includegraphics[width=0.45\linewidth]{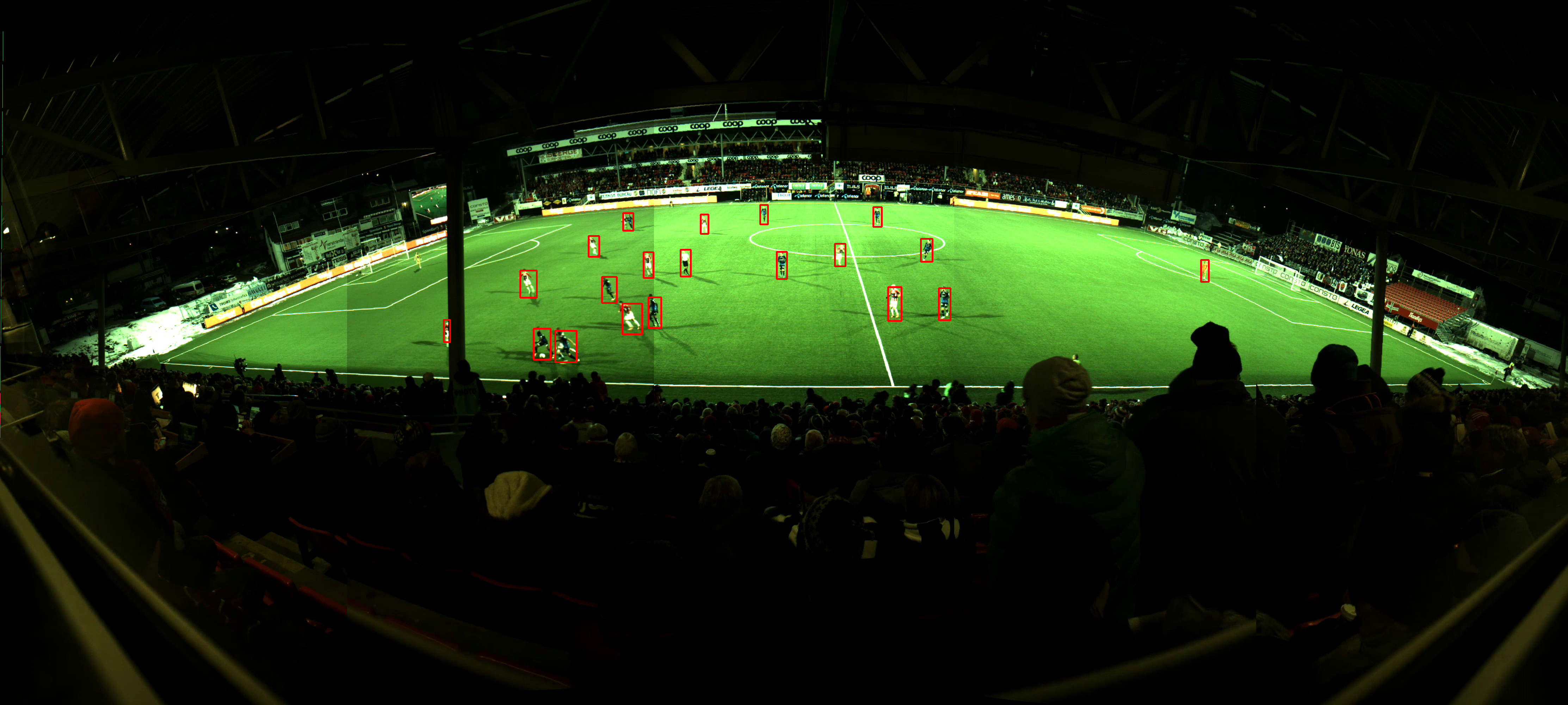}
\includegraphics[width=0.45\linewidth]{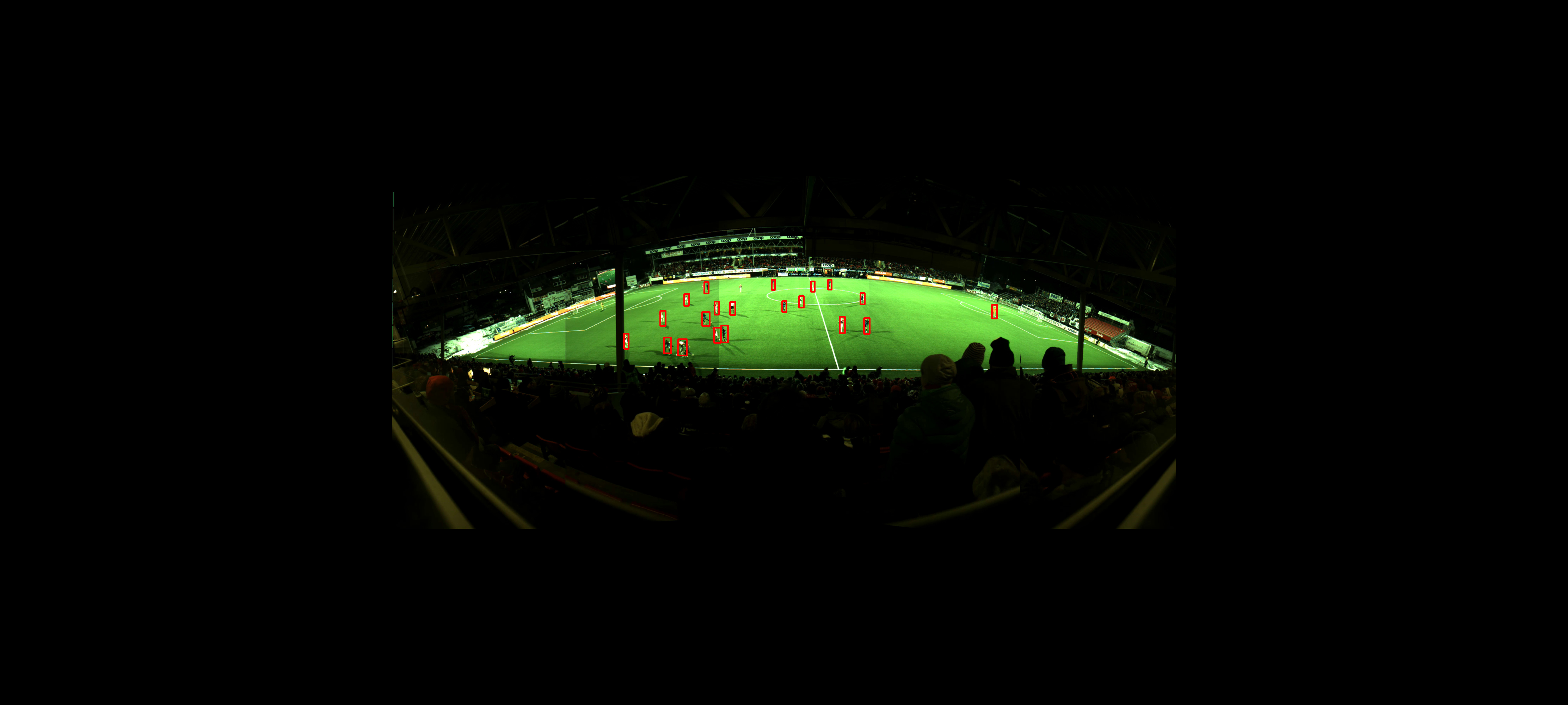}
\caption{images from dataset \cite{panorama_dataset}}
\end{subfigure}
\caption{Player detection results on different datasets, for different down-scaling levels.}
\label{det_res}
\end{figure*}

We show in Figure~\ref{det_res} additional player detection results on the $SPD$ (row 1), $ISSIA$ (row 2) and $TV\_soccer$ (row 3) datasets, as well as on the \textit{panorama} images of \cite{panorama_dataset} (row 4). In order to test the detection on very small players, we down-scale and pad the input images. We show that our network is able to detect almost all the players in  very challenging images, even when a player contains only a few pixels.

\begin{figure}[ht]
\begin{minipage}[b]{0.4\textwidth}
\centering
\includegraphics[width=\linewidth]{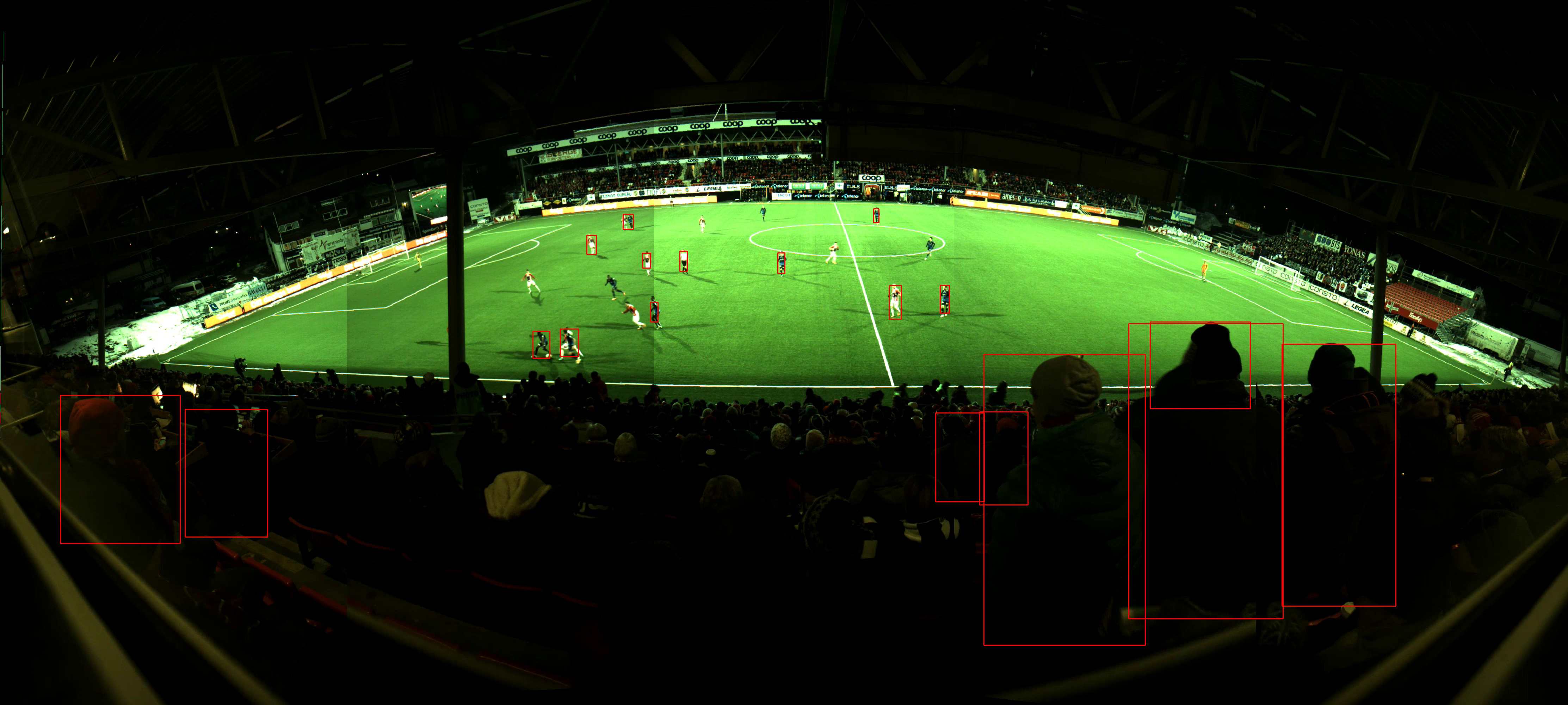}
\end{minipage}
\begin{minipage}[b]{0.4\textwidth}
\centering
\includegraphics[width=\linewidth]{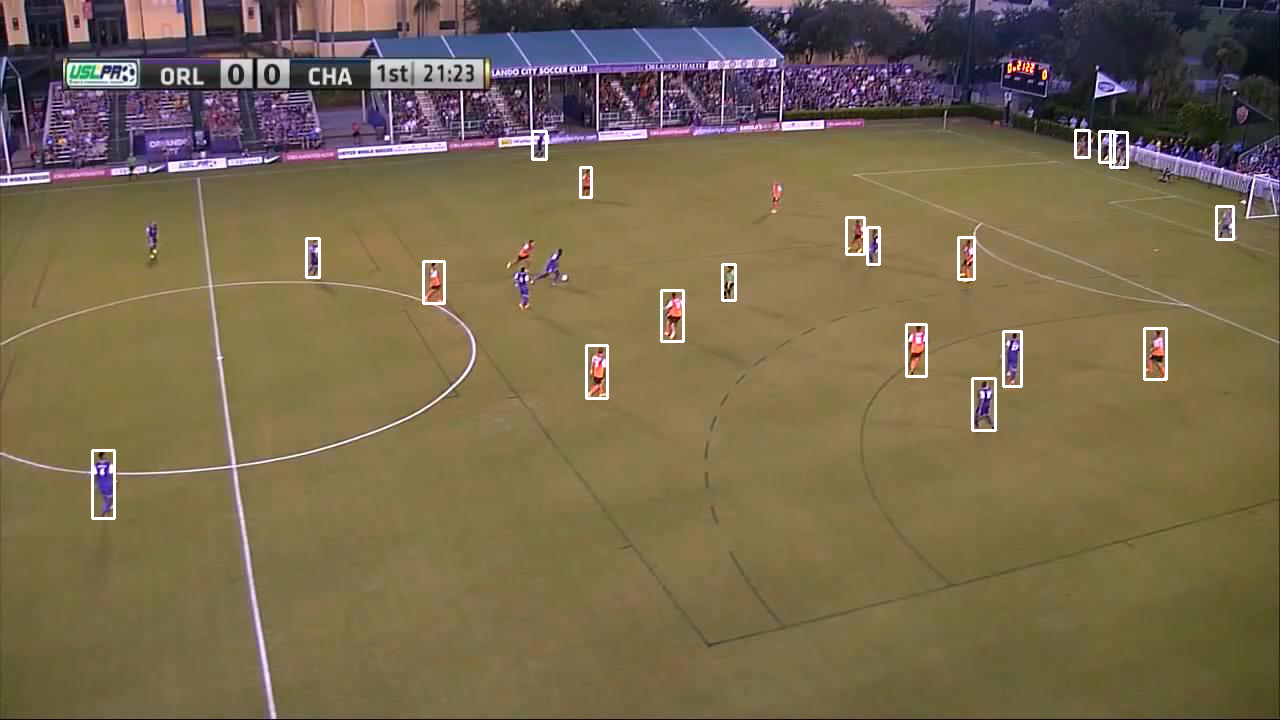}
\end{minipage}
\caption{Player detections results with the human annotations of FPN-FasterRCNN pretrained on COCO ($\mathcal{T}^{init}$).}
\label{pretrained_res}
\end{figure}

To show the strength of our network we present in Figure~\ref{pretrained_res} results with the pretrained FPN-FasterRCNN,  $\mathcal{T}^{init}$, on the same challenging images. It fails at finding small players and detects a lot of non-player humans.

\paragraph{Failure cases} Figure \ref{det_res} includes some failure cases. For example, in the second image of $SPD$ (row 1), non-players are detected on the side of the field. In this particular case, the distinction between players and non-player is challenging because the latter appear on the field. The network can also fail when the background around a player is not green field, like in the third image of $ISSIA$ (row 2). 

\subsection{Player tracking}

\begin{figure*}[ht]
\begin{minipage}[t]{0.35\textwidth}
\centering
\includegraphics[width=\linewidth]{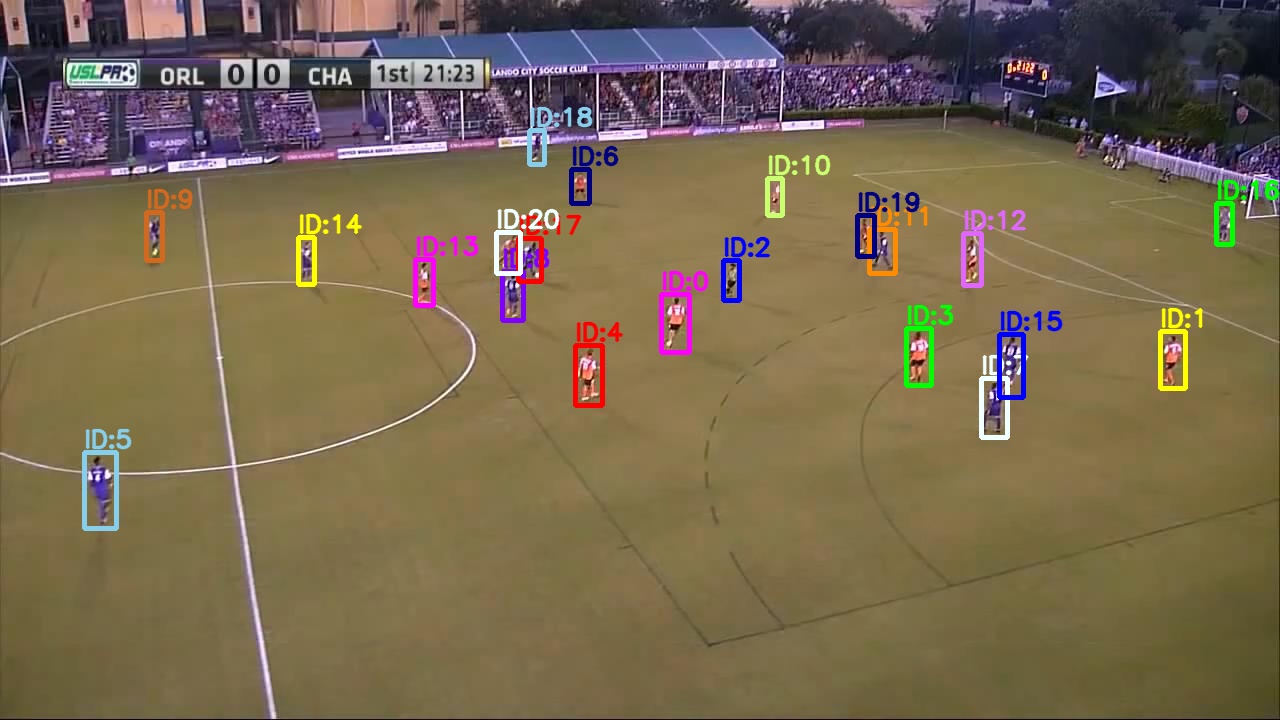}
\includegraphics[width=\linewidth]{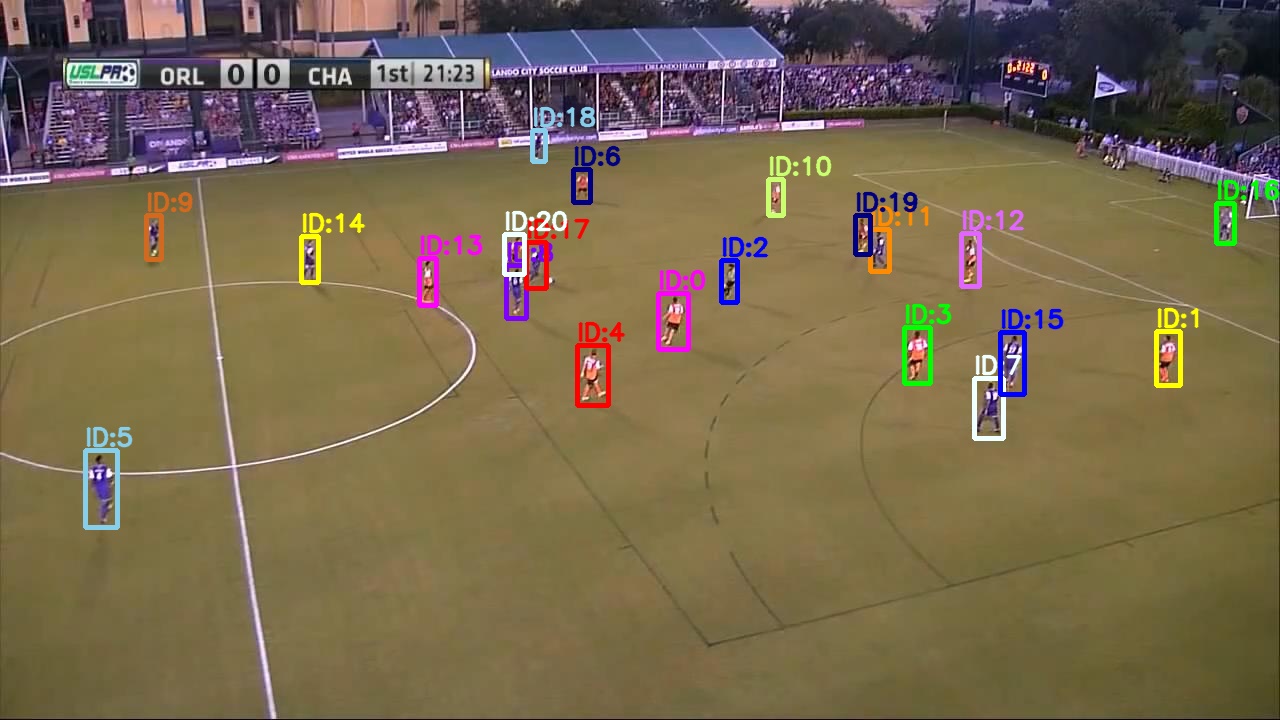}
\includegraphics[width=\linewidth]{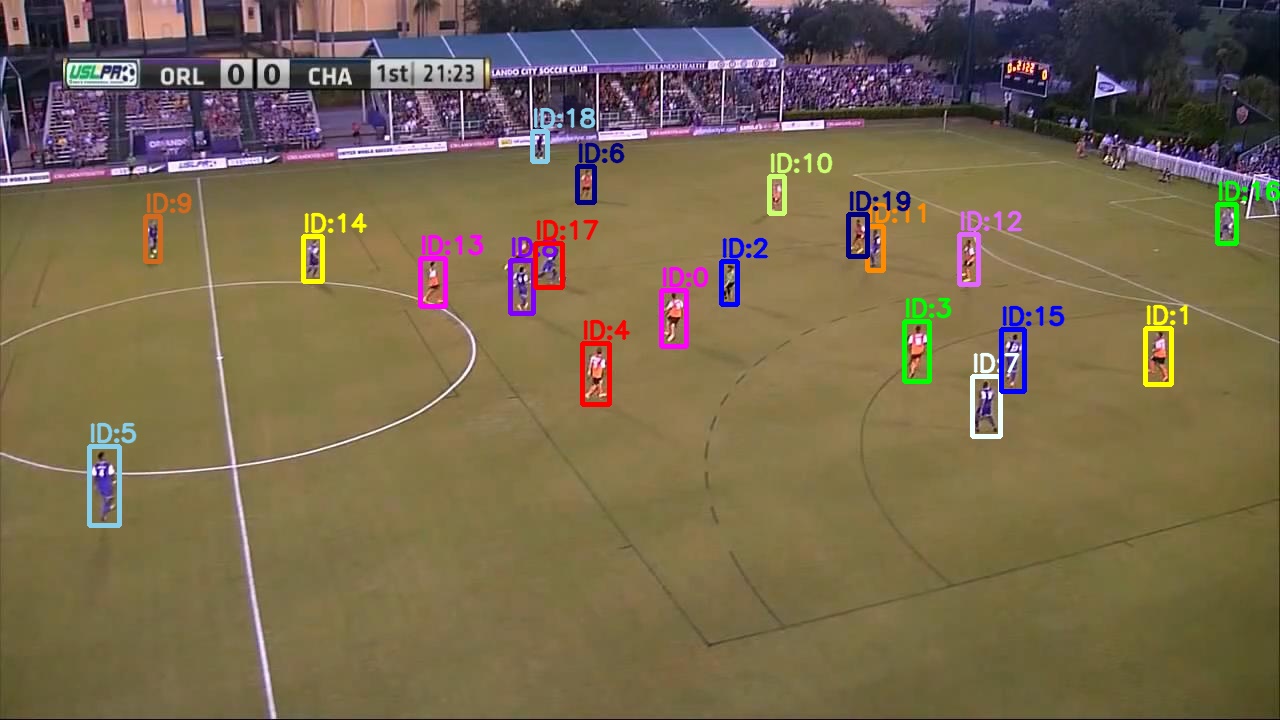}
\includegraphics[width=\linewidth]{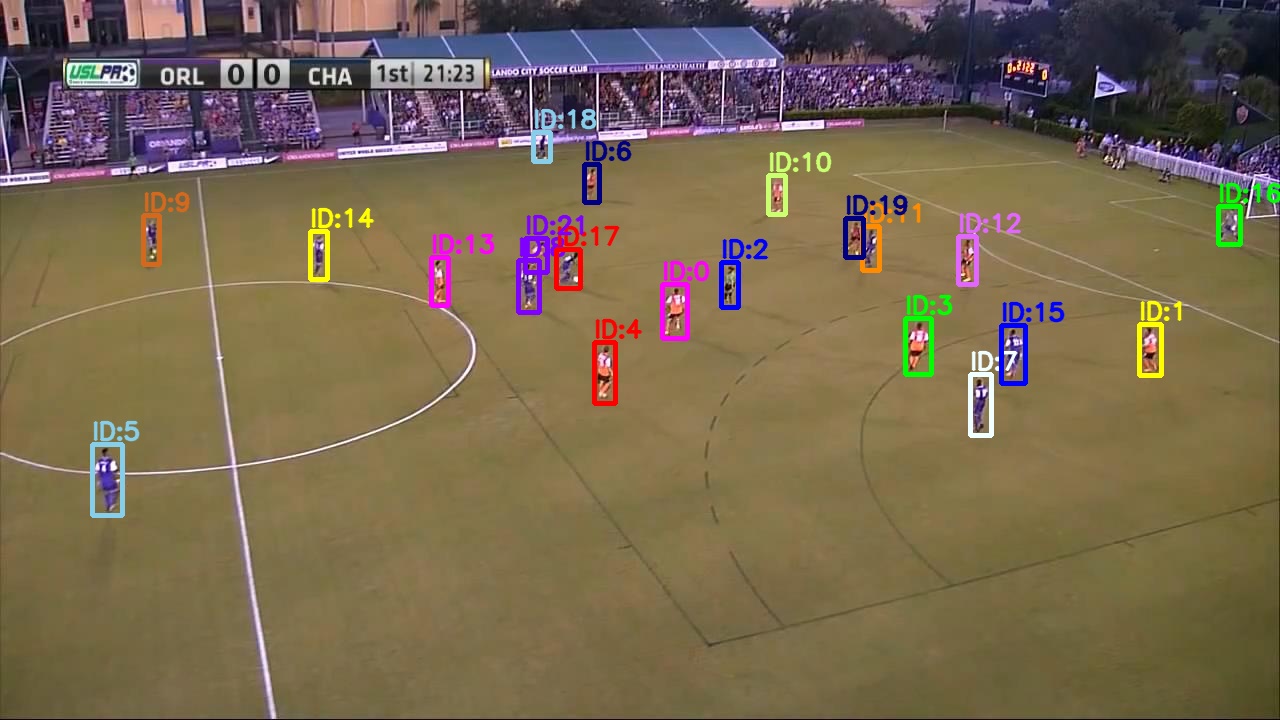}
\\ $SPD$ frames
\end{minipage}
\begin{minipage}[t]{0.5\textwidth}
\centering
\includegraphics[width=\linewidth]{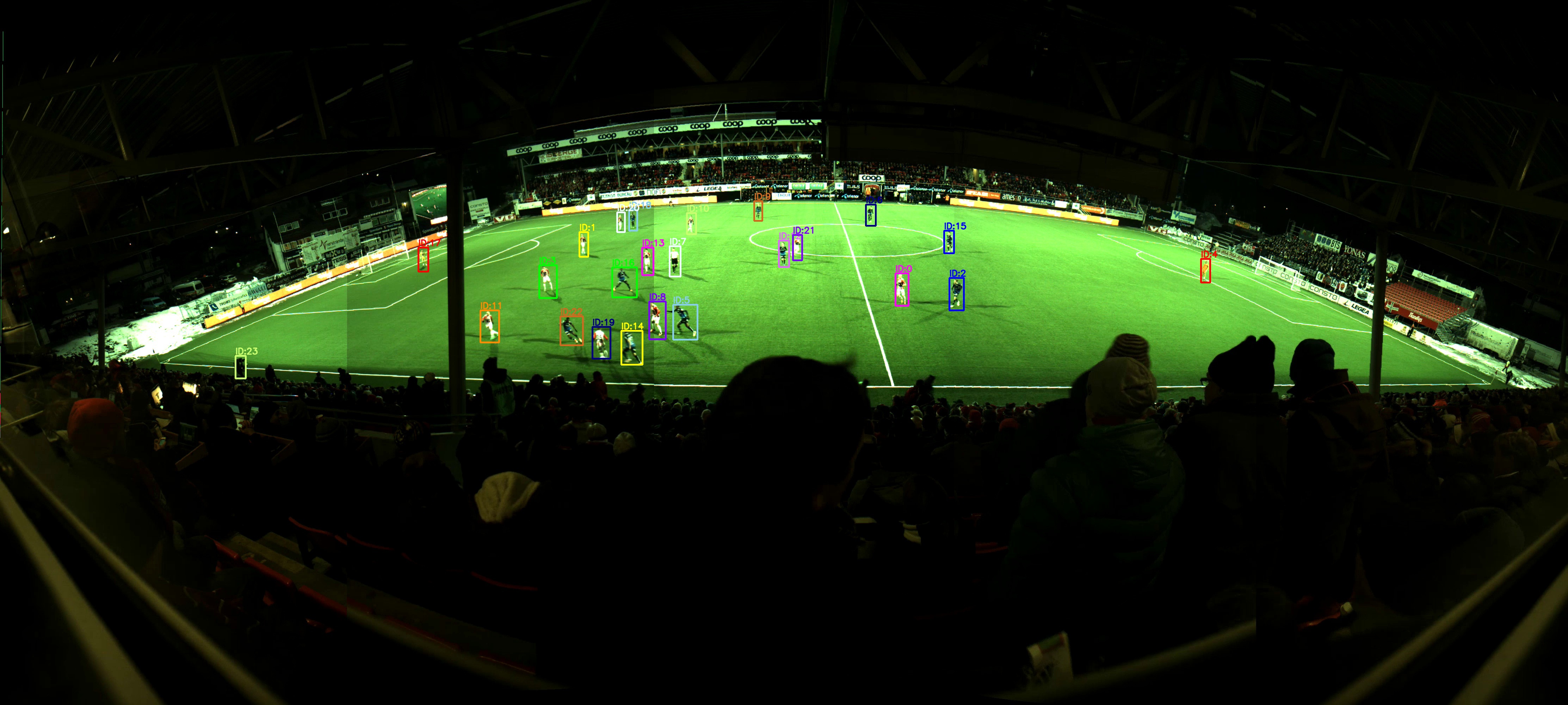}
\includegraphics[width=\linewidth]{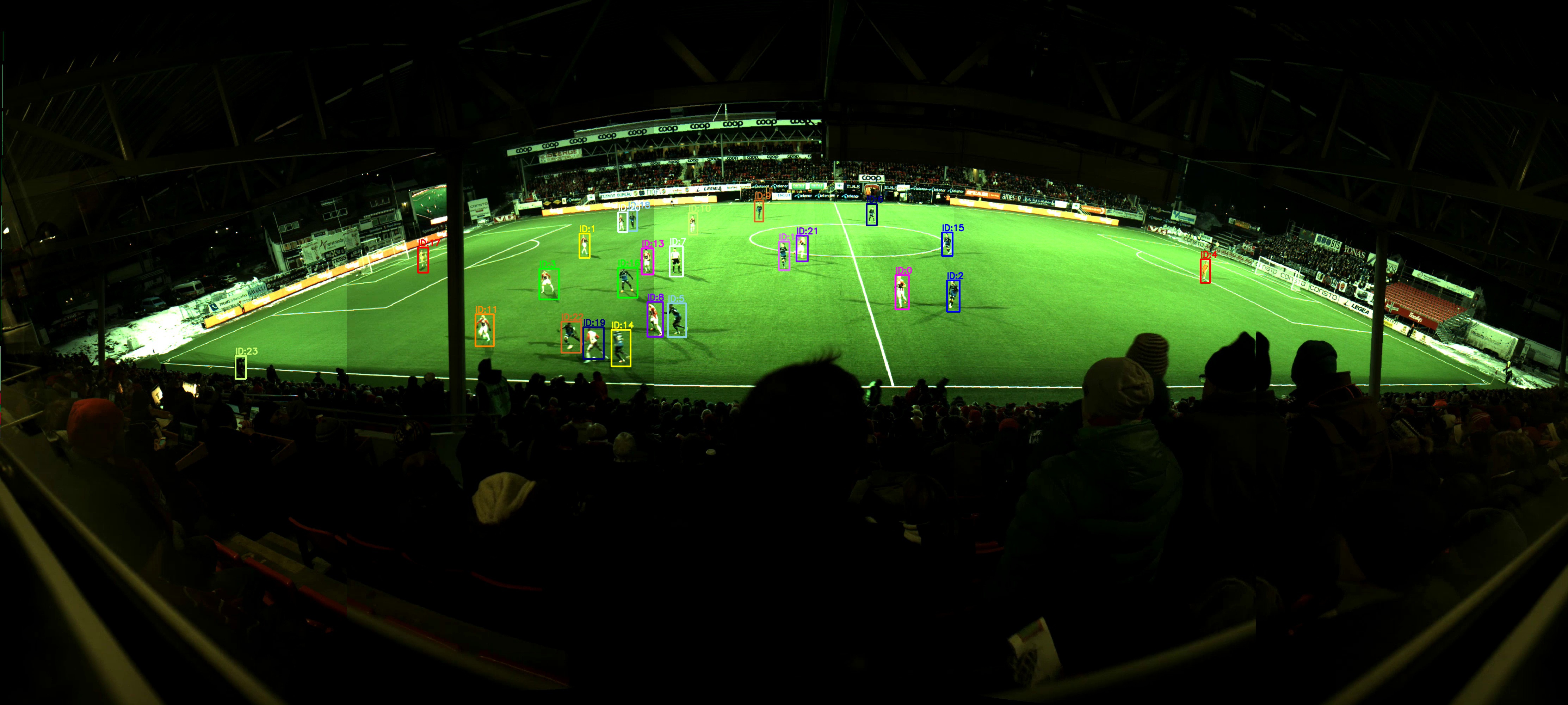}
\includegraphics[width=\linewidth]{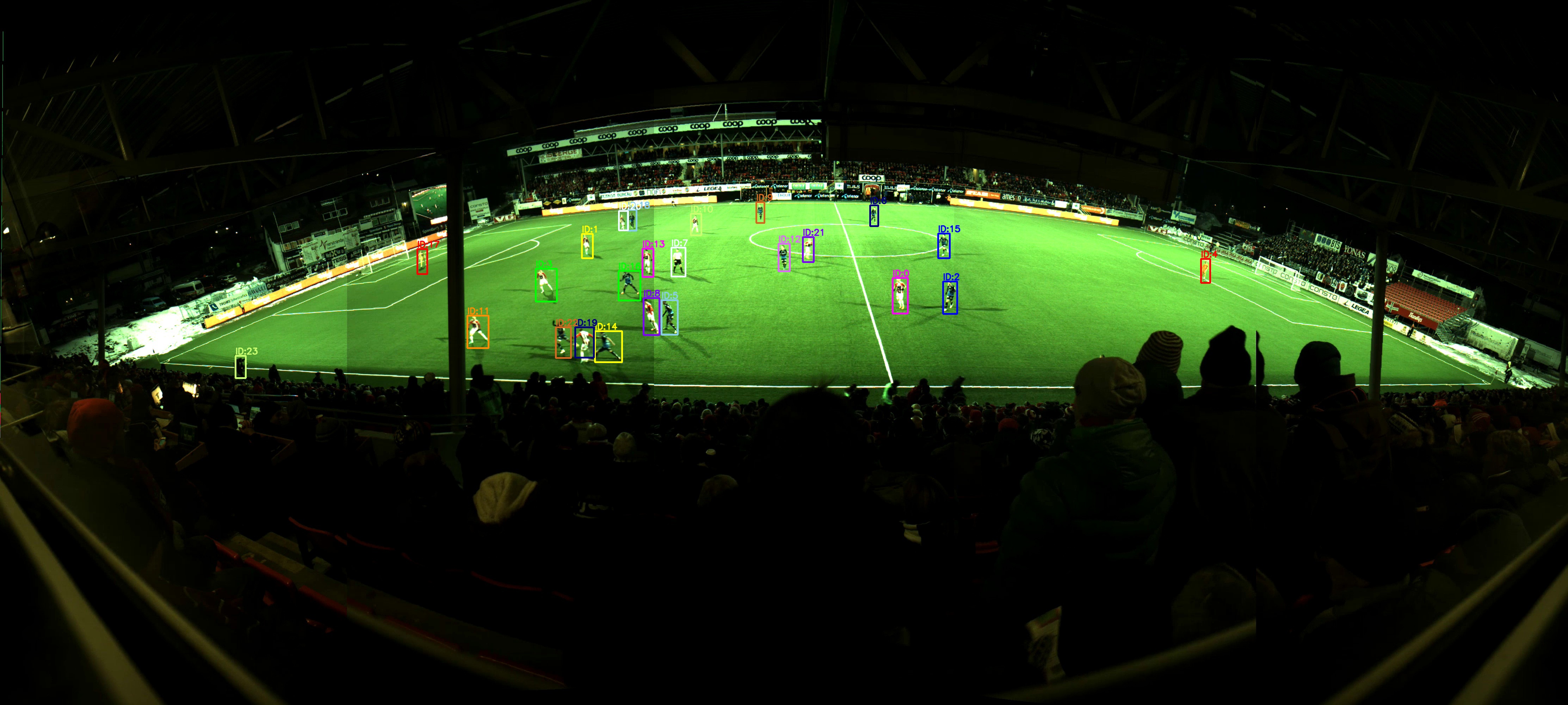}
\includegraphics[width=\linewidth]{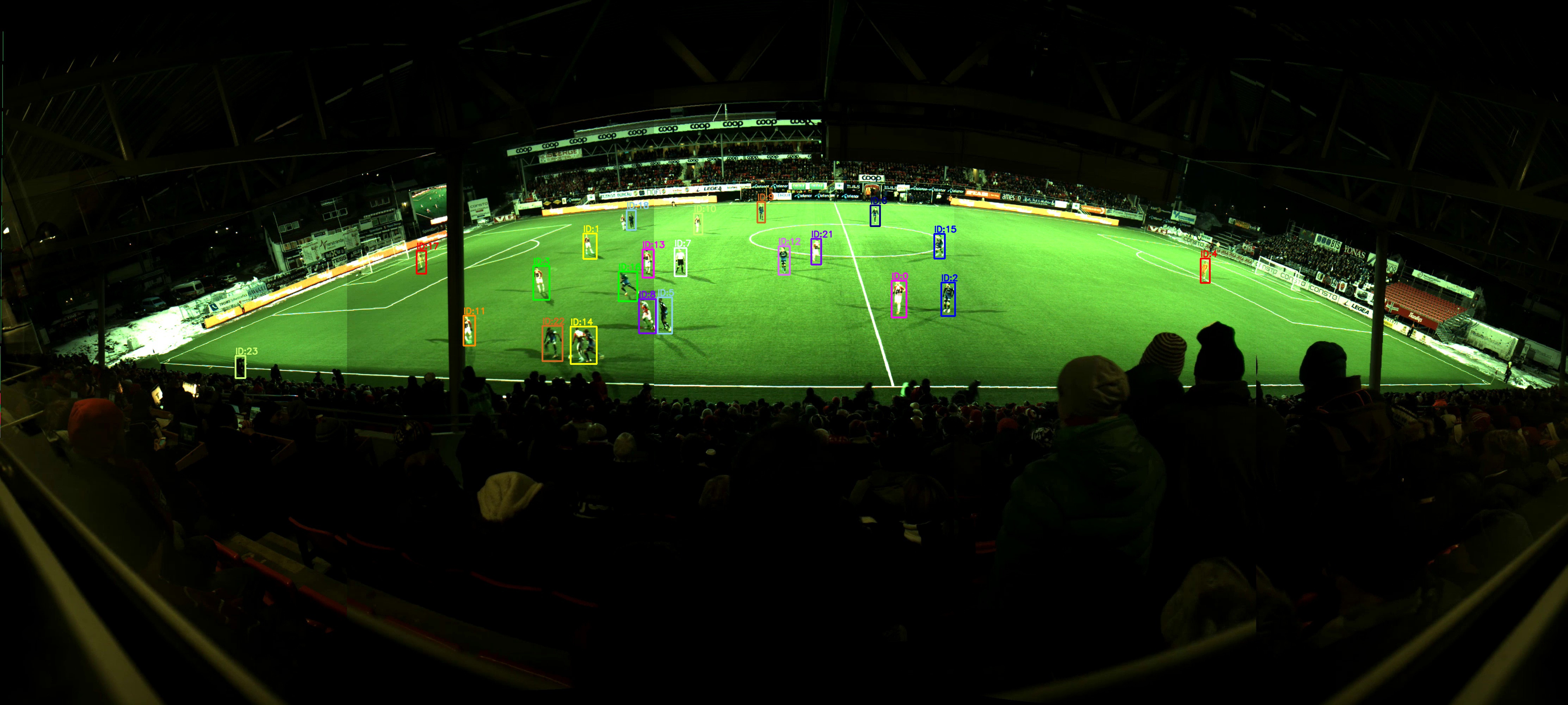}
\\ \textit{panorama} frames
\end{minipage}
\caption{Player tracking results on a $SPD$ sequence and a \textit{panorama} sequence.}
\label{track_res}
\end{figure*}

Figure \ref{track_res} shows two tracking results  on the most challenging $SPD$ and panorama sequences. We present four images of each sequence, one every 5 frames, from top to bottom.

We observe that almost all the players are correctly followed. One failure case appears in the first column, in a typical crowded scene. Due to occlusion, the white player 20 is lost between the frames 2 and 3. The re-ID module fails to re-identify him at frame 4 and it is assigned a new id 21. 
The reason is that the appearance of the player between frame 2 and 4 has changed substantially. Moreover, the bounding box given by the tracker is not really precise for a small player, it can involve only one part of the player, or its close neighbours. In these cases, the visual embedding extracted from this bounding box is not good enough to perform an association. 

\end{document}